\documentclass{article}

\PassOptionsToPackage{numbers, compress}{natbib}

\usepackage[preprint]{neurips_2023}

\usepackage[utf8]{inputenc} 
\usepackage[T1]{fontenc}    
\usepackage{hyperref}       
\usepackage{url}            
\usepackage{booktabs}       
\usepackage{amsfonts}       
\usepackage{amsmath}        
\usepackage{nicefrac}       
\usepackage{microtype}      
\usepackage{xcolor}         
\usepackage{graphicx}
\usepackage{subfigure}
\usepackage{makecell}
\usepackage{bm}
\usepackage{etoc}
\usepackage{cases}

\usepackage[normalem]{ulem}
\usepackage[ruled,vlined]{algorithm2e}

\etocdepthtag.toc{mtchapter}
\etocsettagdepth{mtchapter}{subsection}
\etocsettagdepth{mtappendix}{none}

\title{Formulating Discrete Probability Flow Through Optimal Transport}

\author{%
  Pengze Zhang\thanks{Equal contribution.  This work was done when Pengze Zhang was an intern at WeChat.} \\
  Sun Yat-sen University\\
  \texttt{zhangpz3@mail2.edu.cn} \\
  \And
  Hubery Yin$^*$ \\
  WeChat, Tencent Inc.\\
  \texttt{hubery@tencent.com} \\
  \And
  Chen Li \\
  WeChat, Tencent Inc.\\
  \texttt{chaselli@tencent.com} \\
  \And
  Xiaohua Xie \thanks{Corresponding author.} \\
  Sun Yat-sen University\\
  \texttt{xiexiaoh6@mail.edu.cn} \\}

\begin{document}

\maketitle

\begin{abstract}
  Continuous diffusion models are commonly acknowledged to display a deterministic probability flow, whereas discrete diffusion models do not. In this paper, we aim to establish the fundamental theory for the probability flow of discrete diffusion models. Specifically, we first prove that the continuous probability flow is the Monge optimal transport map under certain conditions, and also present an equivalent evidence for discrete cases.  In view of these findings, we are then able to define the discrete probability flow in line with the principles of optimal transport. Finally, drawing upon our newly established definitions, we propose a novel sampling method that surpasses previous discrete diffusion models in its ability to generate more certain outcomes. Extensive experiments on the synthetic toy dataset and the CIFAR-10 dataset have validated the effectiveness of our proposed discrete probability flow. Code is released at: \url{https://github.com/PangzeCheung/Discrete-Probability-Flow}.
\end{abstract}

\section{Introduction}

The emerging diffusion-based models \cite{sohl2015deep, ho2020denoising,song2019generative, song2020score} have been proven to be an effective technique for modeling data distribution, and generating high-quality texts \cite{NEURIPS2022_1be5bc25,gong2022diffuseq}, images \cite{pmlr-v162-nichol22a, NEURIPS2021_49ad23d1, NEURIPS2022_ec795aea, ramesh2022hierarchical, rombach2022high,JMLR:v23:21-0635} and videos \cite{ho2022video,ho2022imagen,ruan2022mm,NEURIPS2022_94461854,NEURIPS2022_b2fe1ee8}. Considering their generative capabilities have surpassed the previous state-of-the-art results achieved by generative adversarial networks \cite{NEURIPS2021_49ad23d1}, there has been a growing interest in exploring the potential of diffusion models in various advanced applications \cite{9887996,meng2021sdedit,NEURIPS2022_39644677,NEURIPS2022_40e56dab,couairon2023diffedit,liu2023meshdiffusion,tevet2023human,wang2023sketchknitter,hertz2023prompttoprompt,xu2023stochastic}.

Diffusion models are widely recognized for generating samples in a stochastic manner \cite{song2020score}, which complicates the task of defining an encoder that translates a sample to a certain latent space. For instance, by following the configuration proposed by \cite{ho2020denoising}, it has been observed that generated samples from any given initial point have the potential to span the entire support of the data distribution. To achieve a deterministic sampling process while preserving the generative capability, Song \emph{et al.}\cite{song2020score} proposed the probability flow, which provides a deterministic map between the data space and the latent space for continuous diffusion models. Unfortunately,  the situation differs when it comes to discrete models. For instance, considering two binary distributions $(P_0 = \frac{1}{2}, P_1 = \frac{1}{2})$ and $(P_0 = \frac{1}{3}, P_1 = \frac{2}{3})$, there is no deterministic map that can transform the former distribution to the latter one, as it would simply be a permutation. Although some previous research has been conducted on discrete diffusion models with discrete \cite{hoogeboom2021argmax, hoogeboom2021autoregressive, austin2021structured, esser2021imagebart, cohen2022diffusion, johnson2021beyond, gu2022vector} and continuous \cite{campbell2022continuous, sun2022score} time configurations, these works primarily focus on improving the sampling quality and efficiency, while sampling certainty has received less attention. More specifically, there is a conspicuous absence of existing literature addressing the probability flow in discrete diffusion models.

The aim of this study is to establish the fundamental theory of the probability flow for discrete diffusion models. Our paper contributes in the following ways. Firstly, we provide proof that under some conditions the probability flow of continuous diffusion coincides with the Monge optimal transport map during any finite time interval within the range of $\left( 0, \infty \right)$. Secondly, we propose a discrete analogue of the probability flow under the framework of optimal transport, which we have defined as the \emph{discrete probability flow}. Additionally, we identify several properties that are shared by both the continuous and discrete probability flow. Lastly, we propose a novel sampling method based on the aforementioned observations, and we demonstrate its effectiveness in significantly improving the certainty of the sampling outcomes on both synthetic toy dataset and CIFAR-10 dataset.

Proofs for all Propositions are given in the Appendix. For consistency, the probability flow and infinitesimal transport of a process $X_t$ is signified by $\hat{X}_t$ and $\tilde{X}_t$ respectively.

\section{Background on Diffusion Models and Optimal Transport}
\label{backgrounds}

First of all, we review some important concepts from the theory of diffusion models, optimal transport and gradient flow.

\subsection{Continuous state diffusion models}

Diffusion models are generative models that consist of a forward process and a backward process. The forward process transforms the data distribution $p_{data}(x_0)$ into a tractable reference distribution $p_T(x_T)$. The backward process then generates samples from the initial points drawn from $p_T(x_T)$. According to \cite{karras2022elucidating}, the forward process is modeled as the (time-dependent) Ornstein-Uhlenbeck (OU) process:
\begin{equation}
  d X_t = -\theta_t X_t dt + \sigma_t d B_t,
\end{equation}
where $\theta_t \ge 0, \sigma_t > 0, \forall t \geq 0$ and $B_t$ is the Brownian Motion (BM). The backward process is the reverse-time process of the forward process \cite{anderson1982reverse}:
\begin{equation} \label{orig_conti_rev}
  d X_t = [- \theta_t X_t - \sigma_t^2 \nabla_{X_t} \log p(X_t, t)]dt + \sigma_t d \tilde{B}_t,
\end{equation}
where $\tilde{B}_t$ is the reverse-time Brownian motion and $p(X_t, t)$ is the single-time marginal distribution of the forward process, which also serves as the solution to the Fokker-Planck equation \cite{oksendal2013stochastic}:
\begin{equation}
  \frac{\partial}{\partial t} p(x, t) = \theta_t \nabla_x (xp(x, t)) + \frac{1}{2} \sigma_t^2 \Delta_x p(x, t).
\end{equation}

In order to train a diffusion model,  the primary objective is to minimize the discrepancy between the model output $s_\theta (x_t, t)$ and the Stein score function $s (x_t, t) = \nabla_{x_t} \log p(x_t, t)$ \cite{hyvarinen2005estimation}. Song \emph{et al.} \cite{song2019generative} demonstrate that, it is equivalent to match $s_\theta (x_t, t)$ with the conditional score function:
\begin{eqnarray}
  \theta^* = \mathop{\arg\min}\limits_{\theta} \mathbb{E}_t \left\{ \lambda_t \mathbb{E}_{x_0, x_t} \left[ \lVert s_\theta(x_t, t) - \nabla_{x_t} \log p(x_t, t | x_0, 0) \rVert^2 \right]\right\},
\end{eqnarray}
where $\lambda_t$ is a weighting function, $t$ is uniformly sampled over $\left[0, T\right]$ and $p(x_t, t | x_0, 0)$ is the forward conditional distribution.

It is noted that every Ornstein-Uhlenbeck process has an associated probability flow, which is a deterministic process that shares the same single-time marginal distribution \cite{song2020score}. The probability flow is governed by the following Ordinary Differential Equation (ODE):
\begin{equation}
  d \hat{X}_t = [-\theta_t \hat{X}_t - \frac{1}{2} \sigma_t^2 s(\hat{X}_t, t)]dt.
\end{equation}
In accordance with the global version of Picard-Lindelöf theorem \cite{amann2011ordinary} and the adjoint method\cite{pontrjagin1962mathematical, chen2018neural}, the map 
\begin{equation}
\begin{aligned}
  T_{s,t}: \mathbb{R}^n &\longrightarrow \mathbb{R}^n, \\
  \hat{X}_s &\longmapsto \hat{X}_t.
\end{aligned}
\end{equation}
is a diffeomorphism $\forall t \geq s > 0$. The diffeomorphism naturally gives a transport map.

\subsection{Discrete state diffusion models}
In the realm of discrete state diffusion models, there are two primary classifications: the Discrete Time Discrete State (DTDS) models and the Continuous Time Discrete State (CTDS) models, which are founded on Discrete Time Markov Chains (DTMC) and Continuous Time Markov Chains (CTMC), correspondingly. Campbell \emph{et al.}\cite{campbell2022continuous} conducted a comparative analysis of these models and determined that CTDS outperforms DTDS. The DTDS models construct the forward process through the utilization of the conditional distribution $q_{t+1 | t} (x_{t+1} | x_t)$ and employ a neural network to approximate the reverse conditional distribution $q_{t | t+1} (x_t | x_{t+1}) = \frac{q_{t+1 | t} (x_{t+1} | x_t) q_t(x_t)}{q_{t+1}(x_{t+1})}$. In practical applications, it is preferable to parameterize this model using $p^\theta_{0 | t+1}$ \cite{hoogeboom2021argmax, austin2021structured} and obtain $p^\theta_{k | k + 1}$ through
\begin{equation}
\begin{split}
  p^\theta_{k | k+1} (x_k | x_{k+1}) & = \sum_{x_0} q_{k | k+1, 0} (x_k | x_{k+1}, x_0) p^\theta_{0 | k+1} (x_0 | x_{k+1})\\
  & = \sum_{x_0} q_{k+1 | k} (x_{k+1} | x_k) \frac{q_{k | 0}(x_k | x_0)}{q_{k+1 | 0}(x_{k+1} | x_0)} p^\theta_{0 | k+1} (x_0 | x_{k+1}).
\end{split}
\end{equation}
In contrast to DTDS models, a CTDS model is characterized by the (infinitesimal) generator \cite{anderson2012continuous}, or transition rate, $Q_t(x, y)$. The Kolmogorov forward equation \cite{feller2015theory} is:
\begin{equation}
  \frac{\partial}{\partial t} q_{t | s}(x_t | x_s) = \sum_y q_{t|s}(y | x_s) Q_t(y, x_t).
\end{equation}
The reverse process is:
\begin{equation}
  \frac{\partial}{\partial s} q_{s | t}(x_s | x_t) = \sum_y q_{s|t}(y | x_t) R_t(y, x_s).
\end{equation}
The generator of the reverse process can be written by \cite{campbell2022continuous, sun2022score}:
\begin{equation} \label{orig_dis_rev}
  R_t(y, x) = \frac{q_t(x)}{q_t(y)} Q_t(x, y) = \sum_{y_0} \frac{q_{t|0} (x | y_0)}{q_{t|0}(y | y_0)} q_{0 | t} (y_0 | y) Q_t(x, y).
\end{equation}
There are various approaches to train the model, such as the Evidence Lower Bound (ELBO) technique \cite{campbell2022continuous}, and the score-based approach \cite{sun2022score}. It has been observed that the reverse generator can be factorized over dimensions, allowing parallel sampling for each dimension during the reverse process. However, it is important to note that this independence is only possible when the time interval for each step is small.

\subsection{Optimal transport}
The \emph{optimal transport problem} can be formulated in two primary ways, namely the Monge formulation and the Kantorovich formulation \cite{santambrogio2015optimal}. Suppose there are two probability measures $\mu$ and $\nu$ on $(\mathbb{R}^n, \mathcal{B})$, and a cost function $c : \mathbb{R}^n \times \mathbb{R}^n \rightarrow \left[ 0, + \infty \right] $. The \emph{Monge problem} is
\begin{equation}
  \text{(MP)  } \inf_{\text{T}} \left\{ \int c(x, \text{T}(x)) \,{\rm d}\mu(x) : \text{T}_{\texttt{\#}}\mu = \nu \right\}.
\end{equation}
The measure $\text{T}_{\texttt{\#}}\mu$ is defined through $\text{T}_{\texttt{\#}}\mu(A) = \mu(\text{T}^{-1} (A))$ for every $A \in \mathcal{B}$ and is called the \emph{pushforward} of $\mu$ through T. 

It is evident that the Monge Problem (MP) transports the entire mass from a particular point, denoted as $x$, to a single point $\text{T}(x)$. In contrast, Kantorovich provided a more general formulation, referred to as the \emph{Kantorovich problem}:
\begin{equation}
  \text{(KP)  } \inf_{\gamma} \left\{ \int_{\mathbb{R}^n \times \mathbb{R}^n} c\,{\rm d}\gamma : \gamma \in \mit\Pi(\mu, \nu) \right\},
\end{equation}
where $\mit\Pi(\mu, \nu)$ is the set of \emph{transport plans}, i.e.,
\begin{equation}
  \mit\Pi(\mu, \nu) = \left\{ \gamma \in \mathcal{P} (\mathbb{R}^n \times \mathbb{R}^n) : (\pi_x)_{\texttt{\#}}\gamma = \mu, (\pi_y)_{\texttt{\#}}\gamma = \nu \right\},
\end{equation}
where $\pi_x$ and $\pi_y$ are the two projections of $\mathbb{R}^n \times \mathbb{R}^n$ onto $\mathbb{R}^n$.
For measures absolutely continuous with respect to the Lebesgue measure, these two problems are equivalent \cite{villani2009optimal}. However, when the measures are discrete, they are entirely distinct as the constraint of the Monge Problem may never be fulfilled.

\subsection{Fokker-Planck equation by gradient flow}
According to \cite{jordan1998variational}, the Fokker-Planck equation represents the gradient flow of a functional in a metric space. In particular, for Brownian motion, its Fokker-Planck equation, which is also known as the heat diffusion equation, can be expressed as:
\begin{equation} \label{heat_diff_fpe}
  \frac{\partial}{\partial t} p(x, t) = \frac{1}{2} \Delta p(x, t),
\end{equation}
and it represents the gradient flow of the Gibbs-Boltzmann entropy multiplied by $-\frac{1}{2}$:
\begin{equation}
  -\frac{1}{2} S(p) = \frac{1}{2} \int_{\mathbb{R}^n} p(x) \log p(x) \,{\rm d}x.
\label{eq:14}
\end{equation}
It is worth noting that Eq. \ref{eq:14} is the gradient flow of Eq. \ref{heat_diff_fpe} under the 
\emph{2-wasserstein metric} ($W_2$).


Chow \emph{et al.} \cite{chow2012fokker} have developed an analogue in the discrete setting by introducing the discrete Gibbs-Boltzmann entropy:
\begin{eqnarray}
  S(p) = \sum_i p_i~log~p_i,
\end{eqnarray}
and deriving the gradient flow using a newly defined metric (Definition 1 in \cite{chow2012fokker}). Since the discrete model is defined on graph $G(V, E)$, where $V = \{a_1,...,a_N\}$ is the set of vertices, and $E$ is the set of edges, the discrete Fokker-Planck equation with a constant potential can be written as:
\begin{equation} \label{chow_fpe}
  \frac{d}{d t} p_i = \sum_{j \in N(i)} p_j - p_i,
\end{equation}
where $N(i) = \{j \in \{1,2,...,N\}|\{a_i, a_j\}\in E\}$ represents the one-ring neighborhood.

\section{Continuous probability flow}
\label{Continuous}

\subsection{The equivalence of Ornstein-Uhlenbeck processes and Brownian motion}

The diffusion models that are commonly utilized in machine learning are founded on Ornstein-Uhlenbeck processes. First of all, we demonstrate that it is feasible to deterministically convert a time-dependent Ornstein-Uhlenbeck process into a standard Brownian motion.

\paragraph{Proposition 1.} \textit{ Let $X_t$ and $Y_t$ be a time-dependent Ornstein-Uhlenbeck process and a Brownian motion respectively: $dX_t = -\theta_t X_t dt + \sigma_t dB^{(1)}_t $, $dY_t = dB^{(2)}_t$, where $B^{(1)}_t$ and $B^{(2)}_t$ are two independent Brownian motions and $\theta_t \ge 0, \sigma_t > 0,  \forall t \geq 0$. Let $\phi_t = \exp(\int_0^t \theta_\tau\,{\rm d}\tau)$, $\beta_t = \int_0^t (\sigma_\tau \phi_\tau)^2 \,{\rm d}\tau$. Then $X_t$ coincides in law with $\phi^{-1}_{t} Y_{\beta_t}$.}

Building upon the aforementioned proposition, the primary focus of this paper is centered around the standard Brownian motion $dY_t = dB_t$.

\subsection{Probability flow is a Monge map}

Khrulkov \emph{et al.} \cite{khrulkov2023understanding} have proposed a conjecture that the probability flow of Ornstein-Uhlenbeck process is a Monge map. However, they only provided a proof for a simplified case. We demonstrate that under some conditions, the conjecture is correct.

It is important to highlight that the continuous optimal transports presented in this paper are defined exclusively with the cost function: $c(x, y) = \frac{1}{2} | x - y |^2$.

Within the context of generative models, a collection of training samples denoted as $\{ x_i \}_{i=1}^N$ is typically provided, and these samples are intrinsically defined by a distribution:
\begin{equation}\label{init_dist}
  p(x, 0) = \frac{1}{N} \sum_{i=1}^N \delta (x - x_i),
\end{equation}
where $\delta(x)$ represents the Dirac delta function. Given a Brownian motion with an initial distribution in the form of Equation (\ref{init_dist}), the single-time marginal distribution is \cite{oksendal2013stochastic}
\begin{equation}
  p_B(x, t) = \frac{1}{N} \sum_{i=1}^N (2 \pi t)^{-\frac{n}{2}} \exp (- \frac{\left| x - x_i \right| ^ 2}{2t}).
\end{equation}
The probability flow is defined as \cite{song2020score}:
\begin{equation} \label{conti_prob_flow}
  d\hat{Y}_t = - \frac{1}{2} \nabla_{\hat{Y}_t} \log p_B(\hat{Y}_t, t) dt.
\end{equation}
According to \cite{amann2011ordinary, pontrjagin1962mathematical, chen2018neural}, the solution exists for all $t > 0$ and the map $\hat{Y}_{t+s}(\hat{Y}_t)$ is a diffeomorphism for all $t > 0, s \geq 0$. We have discovered that $\hat{Y}_{t+s}(\hat{Y}_t)$ is the Monge map under some conditions and the time does not reach $0$ or $+\infty$.

\paragraph{Proposition 2.}\textit{Given that $Y_0$ follows the initial condition (\ref{init_dist}), and all $x_i$s lie on the same line, the diffeomorphism $\hat{Y}_{t+s}(\hat{Y}_t)$ is the Monge optimal transport map between $p_B(x, t)$ and $p_B(x, t + s)$, $\forall~t > 0, s \geq 0$.}

 There is a counterexample \cite{lavenant2022flow} to demonstrate that the probability flow map does not necessarily provide optimal transport. It is important to note that their case differs from our assumptions in two ways. Firstly, they consider the limit case of $\hat{Y}_{+\infty}(\hat{Y}_0)$. Secondly, the initial distribution of the counterexample does not conform to the form specified in Equation (\ref{init_dist}). Therefore, their counterexample is not applicable to our situation.

It has been shown that the heat diffusion equation can be regarded as the \emph{gradient flow} of the Gibbs-Boltzmann entropy concerning the $W_2$ \emph{metric} \cite{jordan1998variational}. As $W_2$ is associated with optimal transport, it is reasonable to anticipate that the "infinitesimal transport" $\hat{Y}_{t+dt}(\hat{Y}_t)$ is optimal \cite{khrulkov2023understanding}.

In order to interpret the concept of "infinitesimal transport", we utilize the generator of the process $Y_t$. Let $C_c^2(\mathbb{R}^n)$ denote the set of twice continuously differentiable functions on $\mathbb{R}^n$ with compact support. The generator $A_t$ is defined as follows \cite{oksendal2013stochastic}:
\begin{equation}
  \hat{A}_t f = \lim_{\Delta t \rightarrow 0^+} \frac{f(\hat{Y}_{t+\Delta t}) - f(\hat{Y}_t)}{\Delta t}, \forall f \in C_c^2(\mathbb{R}^n).
\end{equation}
It is straightforward to verify that 
\begin{equation}
  \hat{A}_t = -\frac{1}{2} \nabla_{x} \log{p_B(x, t)^T} \nabla_x.
\end{equation}
We define the "infinitesimal transport" to be the diffeomorphism $\tilde{Y}_{t+s} (\tilde{Y}_t)$ where $\tilde{Y}_{t+s}$ evolves according to the following equation 
\begin{equation} \label{infi_trans}
  d \tilde{Y}_{t+s} = -\frac{1}{2} \nabla_{\tilde{Y}_t} \log{p_B (\tilde{Y}_t (\tilde{Y}_{t + s}), t)} ds,
\end{equation}
with the initial condition $\tilde{Y}_t = \hat{Y}_t$. The generator of $\tilde{Y}_{t+s}$ is 
\begin{equation}
  \tilde{A}_{t+s} = - \frac{1}{2} \nabla_{\tilde{Y}_t} \log{p_B (\tilde{Y}_t (\tilde{Y}_{t + s}), t)} \nabla_x.
\end{equation}

\paragraph{Proposition 3.}\textit{ Given any $t > 0$, there exists a $\delta_t > 0$ s.t. $\forall~ 0 < s < \delta_t$, the diffeomorphism $\tilde{Y}_{t+s}(\tilde{Y}_t)$ with the initial condition $\tilde{Y}_t = \hat{Y}_t$ is the Monge optimal transport map.}

Let us return to the original Ornstein-Uhlenbeck process $X_t$. As it is merely a deterministic transformation of the Brownian motion $Y_t$, we can anticipate that the probability flow of $X_t$, denoted by $\hat{X}_t$, will be a Monge map. In fact, this expectation holds true:

\paragraph{Proposition 4.} \textit{Given that $X_0$ follows the initial condition (\ref{init_dist}), and all $x_i$s lie on the same line, the diffeomorphism $\hat{X}_{t+s}(\hat{X}_t)$ is the Monge optimal transport map for all $t > 0, s \geq 0$. }

\section{Discrete probability flow}
\label{Discrete}

The continuous probability flow is deterministic, which means the "mass" at $\hat{Y}_t$ is entirely transported to $\hat{Y}_{t+s}$ during the time interval $\left[ t, t + s \right]$. However, it is widely acknowledged that for discrete distributions $\mu$ and $\nu$, there may not exist a $\text{T}$ such that $\text{T}_{\texttt{\#}}\mu = \nu$. As a result, discrete diffusions cannot possess a deterministic probability flow. To establish the concept of the \emph{discrete probability flow}, we employ the methodology of optimal transport. First of all, a discrete diffusion model is proposed as an analogue of Brownian motion. Secondly, we modified the forward process to create an optimal transport map, which is used to define the {discrete probability flow}. Finally, a novel sampling technique is introduced, which significantly improves the certainty of the sampling outcomes.

\subsection{Constructing discrete probability flow}

It is demonstrated that the process described by Equation (\ref{chow_fpe}) is a discrete equivalent of the heat diffusion process (\ref{heat_diff_fpe}) \cite{chow2012fokker}. We adopt this process as our discrete diffusion model and represent it in a more comprehensive notation.

The discrete diffusion model has $K$ dimensions and $S$ states. The states are denoted by $i = (i_1, i_2, \dots, i_K)$, where $i_j \in \{ 1, 2, \dots, S \}$. The Kolmogorov forward equation for this process is 
\begin{equation} \label{chow_fpe_comp}
  \frac{d }{dt} {P}^{i}_{j} (t | s) = \sum_{j'} {P}^{i}_{j'} (t | s) {Q_D}_{j}^{j'}(t),
\end{equation}
where ${P}^{i}_{j} (t | s)$ means $P(x_t = j | x_s = i)$ and $Q_D$ is defined as:
\begin{equation} \label{chow_Q}
  {Q_D}^{i}_{j} = \begin{cases}
    1, &d_D (i, j) = 1, \\
    -\sum_{j' \in \{k: d_D(i, k) = 1 \}} {Q_D}^{i}_{j'}, &d_D(i, j) = 0, \\
    0, & otherwise,
  \end{cases}
\end{equation}
where $d_D(i,j) = \sum_{l=1}^{K} \left| i_l - j_l \right|$. 
If we let the solution of the Equation (\ref{chow_fpe_comp}) be denoted by $P_D(t | s)$ and assume an initial condition $P_0$, the single-time marginal distribution can be computed as follows:
\begin{equation}
\label{eq:27}
  {P_D}_{i}(t) = \sum_{j} {P_0}_{j} {Q_D}^{j}_{i}(t | 0).
\end{equation}
It is noteworthy that the process defined by $Q_D$ is not an optimal transport map, as there exist \emph{mutual flows} between the states (i.e., there exists two states $i$, $j$ with $Q^{i}_j>0$ and $Q^{j}_i>0$).  Therefore, we propose a modified version that will be proved to be a solution to the Kantorovich problem, namely, an optimal transport plan. The modified version is defined by the following $Q$:

\begin{equation} \label{ot_Q}
  Q^{i}_{j}(t) = \begin{cases}
    \frac{ReLU({P_D}_{i}(t) - {P_D}_{j}(t))}{{P_D}_{i}(t)}, &d_D (i, j) = 1, \\
    -\sum_{j' \in \{k: d_D(i, k) = 1 \}} {Q}^{i}_{j'}(t), &d_D(i, j) = 0, \\
    0, & otherwise.
  \end{cases}
\end{equation}
where
\begin{equation}
  ReLU(x) = \begin{cases}
    x, &x > 0, \\
    0, &x \leq 0.
  \end{cases}
\end{equation}

In order to avoid singular cases, We define $Q^i_j(t)$ to be $0$ when ${P_D}_{i}(t) = 0$. In fact, it is easy to verify that ${P_D}_i(t) > 0$ for all $t > 0$ , $i \in \left\{ 1, 2, \dots, K \right\}$. We will show that the process defined by $Q$ is equivalent in distribution to the one generated by $Q_D$. 

\paragraph{Proposition 5.} \textit{The processes generated by $Q_D$ and $Q$ have the same single-time marginal distribution $\forall t > 0$.}

\paragraph{Proposition 6.} \textit{ Given any $t > 0$, there exists a $\delta_t > 0$ s.t. $\forall~ 0 < s < \delta_t$, the process generated by $Q$ provides an optimal transport map from $P_D(t)$ to $P_D(t + s)$ under the cost $d_D$.}

Proposition 6 demonstrates that $Q_D$ generates a Kantorovich plan between $P_D(t)$ and $P_D(t + s)$ under a certain cost function. On the other hand, the continuous probability flow is the Monge map between $p_B(x, t)$ and $p_B(x, t + s)$. Therefore, it is reasonable to define the process defined by $Q_D$ as the \emph{discrete probability flow} of the original process defined by $Q$.

Furthermore, the "infinitesimal transport" of the discrete process, which is defined by $\frac{d }{ds} \hat{P}(t+s) = \hat{P}(t+s) Q(t) $, also provides an optimal transport map.

\paragraph{Proposition 7.} \textit{Given any $t > 0$, there exists a $\delta_t > 0$ s.t. $\forall~ 0 < s < \delta_t$, the process above provides an optimal transport map from $\hat{P}(t)$ to $\hat{P}(t+s)$ under the cost $d_D$.}

\subsection{Sampling by discrete probability flow}

In order to train the modified model, we employ  a score-based method described in the Score-based Continuous-time Discrete Diffusion Model (SDDM) \cite{sun2022score}. Specifically, we directly learn the conditional probability $P^\theta({i_l}(t) | \left\{ i_1, \dots, i_{l-1}, i_{l+1}, \dots, i_K \right\} (t))$. According to proposition 5, it follows that $P^\theta = {P^\theta}_D$, and consequently, the training process is identical to that of \cite{sun2022score}. For the sake of brevity, we will employ the notation $P^\theta_{i_l | i \backslash i_l}(t)$ to replace $P^\theta({i_l}(t) | \left\{ i_1, \dots, i_{l-1}, i_{l+1}, \dots, i_K \right\} (t))$.

The generator of the reverse process is
\begin{equation} \label{rev_rate}
  R^{i}_{j}(t) = \begin{cases}
    {ReLU}(\frac{{P^\theta_D}_{j_l | i \backslash i_l}(t)}{{P^\theta_D}_{i_l | i \backslash i_l}(t)} - 1), &d_D (i, j) = 1 ~ \text{and} ~ i_l \neq j_l, \\
    -\sum_{j' \in \{k: d_D(i, k) = 1 \}} {R}^{i}_{j'}(t), & d_D (i, j) = 0, \\
    0, &otherwise.
  \end{cases}
\end{equation}
We use the Euler's method to generate samples. Given the time step length $\epsilon$, the transition probabilities for dimension $l$ is:
\begin{equation} \label{euler}
  P^\theta({i_l}(t - \epsilon) | i(t)) = \begin{cases}
    \epsilon R^{i(t)}_{i_1(t), \dots, i_l(t - \epsilon), \dots, i_k(t)}(t), & i_l(t - \epsilon) \neq i_l(t),\\
    1 + \epsilon R^{i(t)}_{i(t)}(t), & i_l(t - \epsilon) = i_l(t).
  \end{cases}
\end{equation}
When $\epsilon$ is small, the reverse conditional distribution has the factorized probability:
\begin{equation} \label{rev_sample}
  P^\theta(i(t - \epsilon) | i(t)) = {\Pi}_{l=1}^K P^\theta({i_l}(t - \epsilon) | i(t))
\end{equation}

In this way, it becomes possible to generate samples by sequentially sampling from the reverse conditional distribution \ref{rev_sample}.

\paragraph{Transition to higher probability states} The reverse process of the continuous probability flow, as described in Equation (\ref{conti_prob_flow}), causes particles to move towards areas with higher logarithmic probability densities. As the logarithm function is monotonically increasing, this reverse flow pushes particles to higher probability density states. This phenomenon is also observed in the discrete probability flow. By examining the reverse generator, as shown in Equation (\ref{rev_rate}), it can be determined that the transition rate $R^i_j(t) > 0$ only when the destination state $j$ has a higher probability than the source state $i$. This implies that transitions only occur in higher probability states. In contrast, the original continuous reverse process (\ref{orig_conti_rev}) and the discrete reverse process from (\ref{orig_dis_rev}) allow any transitions.

\paragraph{Reduction of Standard Deviation} We measure the certainty of the sampling method by the expectation of the Conditional Standard  Deviation (CSD):
\begin{equation} \label{eq:var}
  CSD_{s,t}(X) = \mathbb{E}_{X_t} [ \text{Std} (X_s | X_t)],
\end{equation}
where $\text{Std}(X_s | X_t) = \text{Var}^{\frac{1}{2}}(X_s | X_t) = \mathbb{E}^{\frac{1}{2}}_{X_s}[X_s - \mathbb{E}_{X_s}[X_s|X_t]| X_t]$. $CSD_{s,t}(X)$ is $0$ when the process is deterministic, such as the continuous probability flow. In the discrete situation, there does not exist any deterministic map. However, our discrete probability flow significantly reduces $CSD_{s,t}(X)$. Table \ref{tab:ecv} presents numerical evidence of this phenomenon. Therefore, we posit that the discrete probability flow enhances the certainty of the sampling outcomes.

\section{Related Work}
The concept of probability flow was initially introduced in \cite{song2020score} as a deterministic alternative to the It\^{o} diffusion. In the work \cite{song2021denoising}, they presented the Denoising Diffusion Implicit Model (DDIM) and demonstrated its equivalence to the probability flow. Subsequently, \cite{khrulkov2023understanding} investigated the relationship between the probability flow and optimal transport. They hypothesized that the probability flow could be considered a Monge optimal transition map and provided a proof for a specific case. Additionally, they conducted numerical experiments that supported their conjecture, showing negligible errors. However, \cite{lavenant2022flow} has discovered an initial distribution that renders probability flow not optimal.

The discrete diffusion models were first introduced by \cite{sohl2015deep}, who considered a binary model. Following the success of continuous diffusion models, discrete models have garnered more attention. The bulk of research on discrete models has focused primarily on the design of the forward process \cite{hoogeboom2021argmax, hoogeboom2021autoregressive, austin2021structured, bond2022unleashing, johnson2021beyond, gu2022vector, cohen2022diffusion}. Continuous time discrete state models were introduced by \cite{campbell2022continuous} and subsequently developed by \cite{sun2022score}.  

\section{Experiments}

We conduct numerical experiments using our novel sampling method by Discrete Probability Flow (DPF) on synthetic data. The primary goal is to demonstrate that our method can generate samples of comparable quality with higher certainty.

\setlength{\tabcolsep}{1.5mm}{
\centering
\begin{table}[!t]
\centering
  \caption{Comparison of generation quality for SDDM and DPF, in terms of MMD with Laplace kernel using bandwith=0.1. Lower values indicate superior quality.}
  \renewcommand\arraystretch{0.8}
  \begin{tabular}{cccccccc}
  \Xhline{1pt}
  \multicolumn{1}{c|}{}     & 2spirals & 8gaussians & checkerboard & circles & moons & pinwheel & swissroll \\ \Xhline{1pt}
                            & \multicolumn{7}{c}{discrete dimension = 32, state size = 2}              \\ \hline
  \multicolumn{1}{c|}{SDDM} & 2.18e-06 & 4.28e-06  & 1.33e-06 & 6.22e-06 & 5.62e-06  & 2.10e-06 & 4.27e-06   \\
  \multicolumn{1}{c|}{DPF (ours)}   & 1.89e-05 & 1.09e-05  & 2.22e-05 & 3.27e-05 & 2.42e-05  & 1.60e-05 & 2.18e-05    \\ \hline
                            & \multicolumn{7}{c}{discrete dimension = 16, state size = 5}              \\ \hline
  \multicolumn{1}{c|}{SDDM} & 2.06e-4         & 1.01e-4           & 2.43e-4  & 1.74e-4             & 2.20e-4        & 3.37e-4      & 1.43e-4         \\
  \multicolumn{1}{c|}{DPF (ours)}   &  3.87e-4        &  5.87e-4          & 4.93e-4       & 3.83e-4      & 3.43e-4        & 6.64e-4      & 3.20e-4         \\ \hline
                            & \multicolumn{7}{c}{discrete dimension = 12, 
  state size = 10}             \\ \hline
  \multicolumn{1}{c|}{SDDM} & 5.52e-4         & 3.01e-4           & 4.39e-4  & 4.22e-4             & 2.71e-4        & 2.90e-4      & 3.39e-4         \\
  \multicolumn{1}{c|}{DPF (ours)}   &  7.19e-4        &  3.49e-4          & 5.99e-4             & 6.65e-4        & 4.34e-4      & 4.14e-4         & 5.17e-4          \\ \Xhline{1pt}
  \end{tabular}
  \label{tab:mmd}
\end{table}
} 

\setlength{\tabcolsep}{1.8mm}{\begin{table}[t!]
\centering
  \caption{Comparison of certainty for SDDM and DPF, in terms of $CSD$ on 4,000 initial points, each of which has 10 generated samples. Lower values indicate superior certainty.}
  \renewcommand\arraystretch{0.8}
  \begin{tabular}{cccccccc}
  \Xhline{1pt}
  \multicolumn{1}{c|}{}     & 2spirals & 8gaussians & checkerboard & circles & moons & pinwheel & swissroll \\ \Xhline{1pt}
                            & \multicolumn{7}{c}{discrete dimension = 32, state size = 2}              \\ \hline
  \multicolumn{1}{c|}{SDDM} & 14.3053 & 14.1882  & 14.7433 & 14.4327 & 14.1739  & 14.0450 & 14.0548   \\
  \multicolumn{1}{c|}{DPF (ours)}   & 2.1719 & 1.7945  & 2.0693 & 1.7210 & 2.0573  & 2.1834 & 1.8892    \\ \hline
                            & \multicolumn{7}{c}{discrete dimension = 16, state size = 5}              \\ \hline
  \multicolumn{1}{c|}{SDDM} & 14.4645         & 14.6143           & 14.6963  & 14.4807             & 14.2397        & 14.2466      & 14.2659         \\
  \multicolumn{1}{c|}{DPF (ours)}   &  1.9711        &  1.9367          & 1.4172       & 1.7185      & 1.7668        & 1.9633      & 1.6665         \\ \hline
                            & \multicolumn{7}{c}{discrete dimension = 12, state size = 10}             \\ \hline
  \multicolumn{1}{c|}{SDDM} & 12.8463         & 12.7933           & 13.0158  & 12.9232             & 12.6665        & 12.7634      & 12.7880         \\
  \multicolumn{1}{c|}{DPF (ours)}   &  1.8123        &  1.3178          & 1.1348             & 1.4625        & 1.4859      & 1.8435         & 1.5227          \\ \Xhline{1pt}
  \end{tabular}
  \label{tab:ecv}
\end{table}
}

\begin{figure*}[!t]
  \centering
  \includegraphics[width=0.8\textwidth]{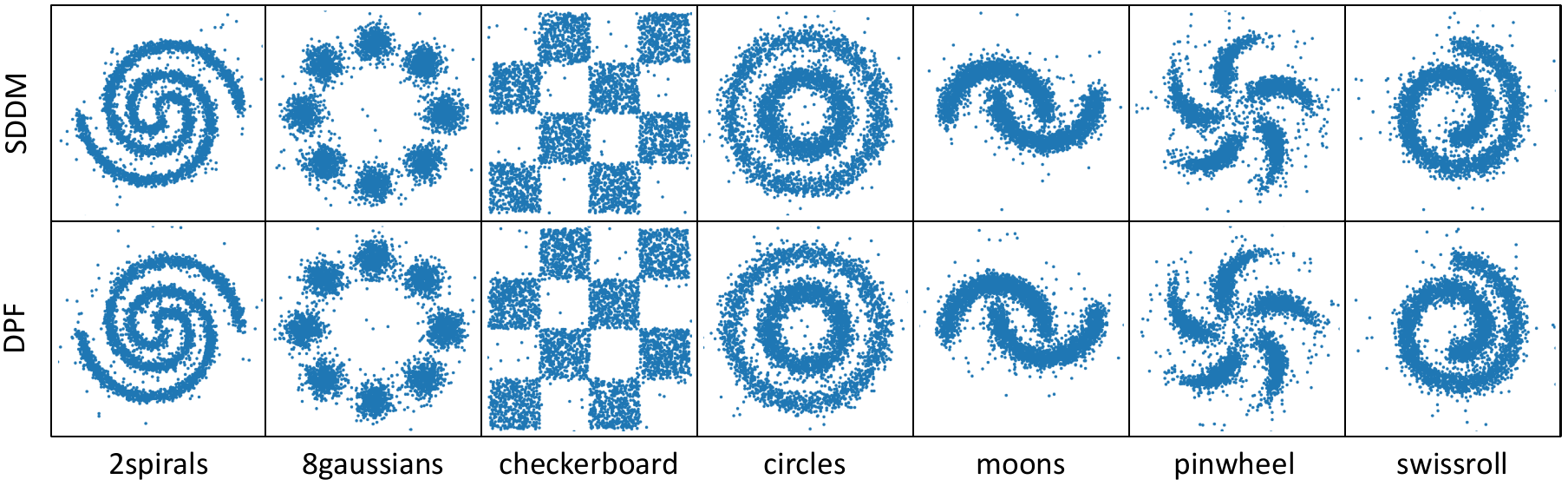} 
  \caption{Visualization of the generation quality on generated binary samples for SDDM and DPF. }
  \label{fig:fig1}
\end{figure*}
\begin{figure*}[t]
  \centering
  \includegraphics[width=0.8\textwidth]{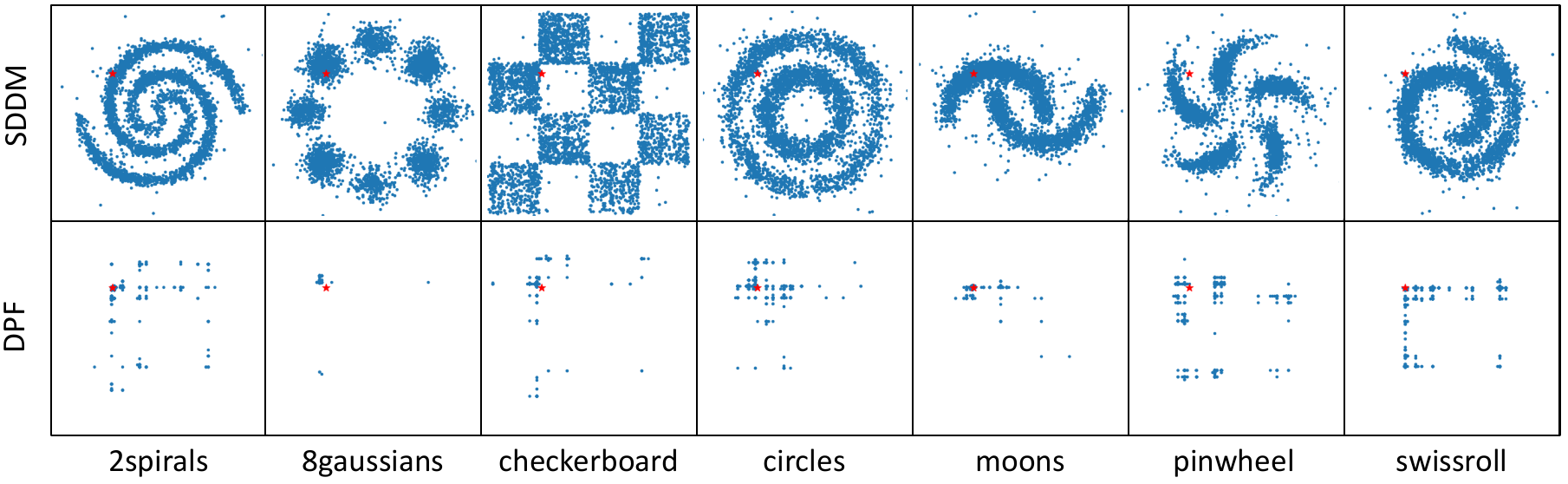} 
  \caption{Visualization of the generating certainty on generated binary samples for SDDM and DPF. All the samples (in blue) are randomly generated from the single initial point (in red).}
  \label{fig:fig2}
\end{figure*}
\begin{figure*}[t!]
  \centering
\includegraphics[width=1\textwidth]{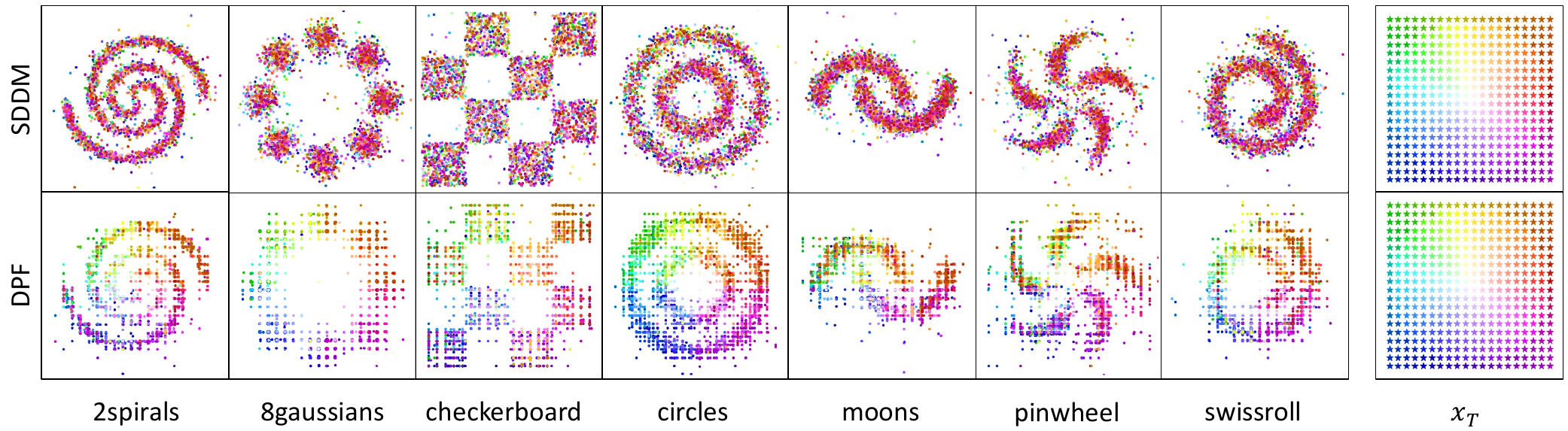} 
  \caption{Visualization of the generated binary samples from the given initial points $\bm{x}_T$. 
  Different colors distinguish the generated samples from different initial points $\bm{x}_T$.}
  \label{fig:fig3}
\end{figure*}

Experiments are conducted on synthetic data  using the same setup as SDDM \cite{sun2022score}, with the exception that we replaced the generator $Q$ with Equation (\ref{chow_Q}). In addition to the binary situation ($S = 2$) studied in \cite{sun2022score}, we also perform experiments on synthetic data with the state size $S$ set to 5 and 10. To evaluate the quality of the generated samples, we generated 40,000 / 4,000  samples for binary data / other type of data using SDDM and DPF, and measured the Maximum Mean Discrepancy (MMD) with the Laplace kernel \cite{gretton2012kernel}. The results are shown in Table \ref{tab:mmd}. It can be seen that the MMD value obtained using DPF is slightly higher than that of SDDM, which may be attributed to the structure of the reverse generator \ref{orig_dis_rev}. Specifically, DPF approximates an additional term, $Q_t(y, x)$, with the neural network, which potentially introduces additional errors to the sampling process, leading to a higher MMD value compared to SDDM. However, such difference is minimal and does not significantly impact the quality of the generated samples. As evident from the visualization of the distributions obtained from SDDM and DPF in Figure \ref{fig:fig1}, it is clear that DPF can generate samples that are comparable to those generated by SDDM.

In addition, we also compare the sampling certainty of DPF and SDDM by computing $CSD_{s,t}$ using a Monte-Carlo based method. Specifically, we set $s = 0$ and $t = T$, and sample 4,000 $x_t$s with 10 $x_s$s for each $x_t$. We then estimate $\mathbb{E}(x_s | x_t)$ and $\text{Std}(x_s | x_t)$ using the sample mean and sample standard deviation, respectively. The results of certainty are presented in Table \ref{tab:ecv}. Our findings indicate that DPF significantly reduces the $CSD$, which suggests a higher certainty. Additionally, we visualize the results of 4,000 generated samples (in blue) from a single initial point (in red) in the binary case in Figure \ref{fig:fig2}. It is apparent that the sampling of SDDM exhibits high uncertainty, as it can sample the entire pattern from a single initial point. In contrast, our method reduces such uncertainty and is only able to sample a limited number of states. 

To provide a more intuitive representation of the generated samples originating from various initial points, we select $20 \times 20$ initial points arranged in the grid, and distinguish them using different colors. Subsequently, we visualize the results by sampling 10 outcomes from each initial point, as shown in Figure \ref{fig:fig3}. We observe that the visualization of SDDM samples appears disorganized, indicating significant uncertainty. In contrast, the visualization of DPF samples exhibits clear regularity, manifesting in two aspects: (1) the generated samples from the same initial point using DPF are clustered by color, demonstrating the better sampling certainty of our DPF. (2) Both of the generated samples and initial points are colored similarly at each position. For example, in the lower right area, a majority of the generated samples are colored purple, which corresponds to the color assigned to the initial points $x_T$ in that area. This observation demonstrates that most of the sampling results obtained through DPF are closer to their respective initial points, aligning with our design intention of optimal transport. It is worth noting that similar phenomena are observed across different state sizes, and we have provided these results in the Appendix.

Finally, we extended our DPF to the CIFAR-10 dataset, and compare it with the $\tau$LDR-0 method proposed in \cite{campbell2022continuous}. The visualization results are shown in Figure \ref{fig:cifar}. It can be seen that our method greatly reduces the uncertainty of generating images by sampling from the same initial $x_T$. Detailed experimental settings and more experimental results are presented in the Appendix.

\section{Discussion}
In this study, we introduce a discrete counterpart of the probability flow and established its connections with the continuous formulations. We began by demonstrating that the continuous probability flow corresponds to a Monge optimal transport map. Subsequently, we proposed a method to modify a discrete diffusion model to achieve a Kantorovich plan, which naturally defines the discrete probability flow. We also discovered shared properties between continuous and discrete probability flows. Finally, we propose a novel sampling method that significantly reduces sampling uncertainty. However, there are still remaining aspects to be explored in the context of the discrete probability flow. For instance, to obtain more general conclusions under a general initial condition, the semi-group method \cite{yosida1967functional} could be employed. Additionally, while we have proven the existence of a Kantorovich plan in a small time interval, it is possible to extend this to a global solution. Moreover, the definition of the probability flow has been limited to a specific type of discrete diffusion model, which also could be extended to a broader range of models. These topics remain open for future studies.

\section{Acknowledgments and Disclosure of Funding}
We would like to thank all the reviewers for their constructive comments. Our work was supported in National Natural Science Foundation of China (NSFC) under Grant No.U22A2095 and No.62072482.

\begin{figure*}[t]
  \centering
  \includegraphics[width=1\textwidth]{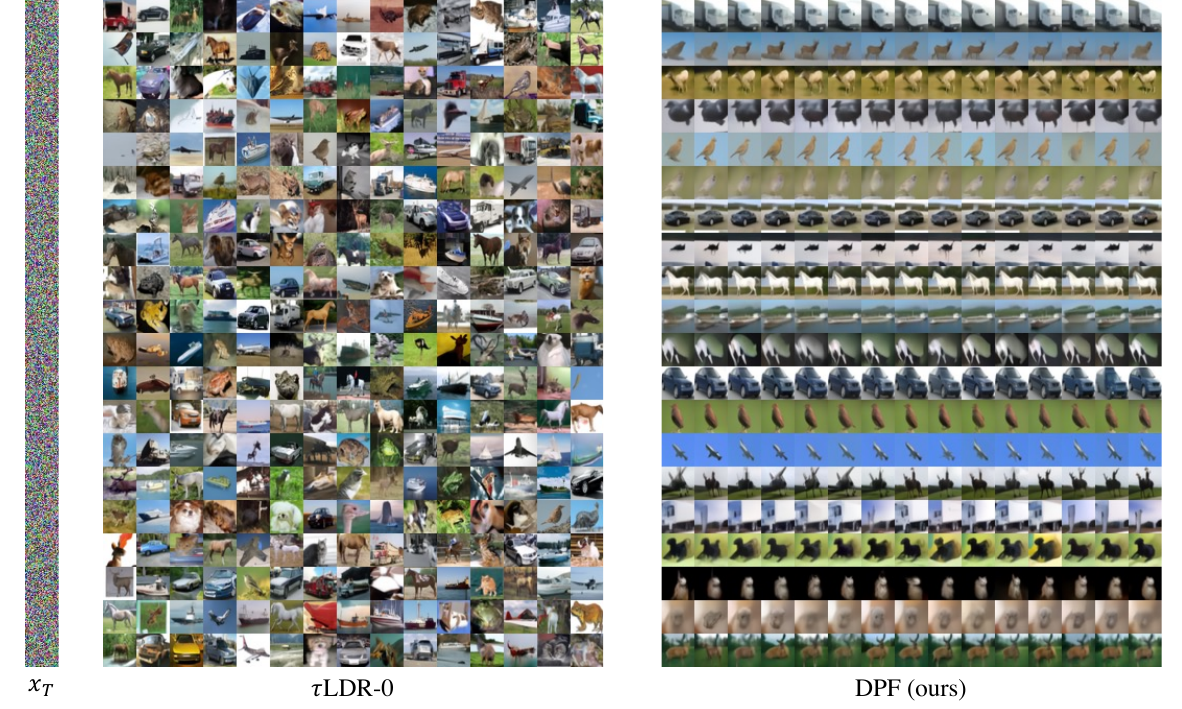} 
  \caption{Image modeling on CIFAR-10 dataset. The figure is divided into three groups: initial points $x_T$, sampling results of $\tau$LDR-0, and sampling results of our DPF. For each row, the sampled images are obtained from the same initial point.}
  \label{fig:cifar}
\end{figure*}

\bibliographystyle{plain}
\bibliography{reference}


\newpage
\appendix

{\huge \bfseries Appendix}
\etocdepthtag.toc{mtappendix}
\etocsettagdepth{mtchapter}{none}
\etocsettagdepth{mtappendix}{subsection}
\tableofcontents

\newcommand{\intd}{\,{\rm d}}
\newcommand{\loggaui}{\frac{|x_t - x_i|^2}{2t}}
\newcommand{\loggauj}{\frac{|x_t - x_j|^2}{2t}}
\newcommand{\diri} {\frac{x_t - x_i}{t}}
\newcommand{\dirj} {\frac{x_t - x_j}{t}}

\section{Overview of our DPF} \label{sec:ApdxPrimerCTMC}
\begin{figure*}[t]
  \centering
  \includegraphics[width=1\textwidth]{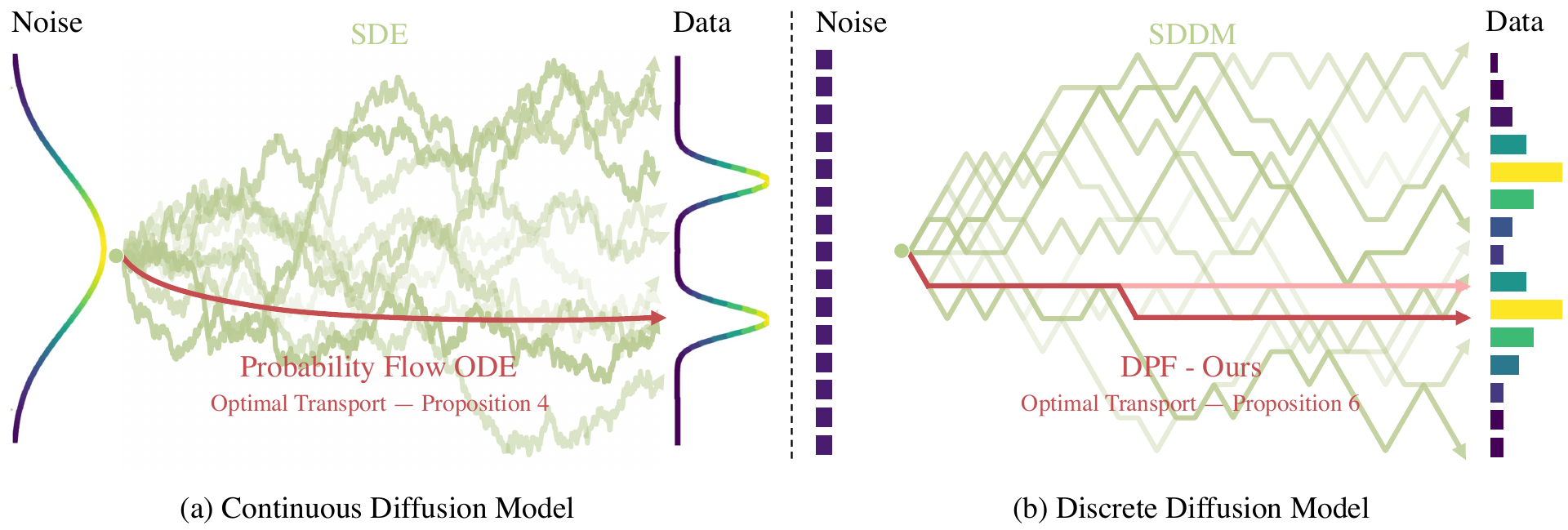} 
  \caption{Schematic representation of different diffusion models. }
  \label{fig:DDPM}
\end{figure*}

To elucidate our methodology more intuitively, we include schematic diagrams in Figure \ref{fig:DDPM}, illustrating the sampling procedure from various diffusion models. Broadly speaking, diffusion models can be classified into two categories based on the nature of the underlying data space: continuous diffusion models and discrete diffusion models. Figure \ref{fig:DDPM} (a) provides an illustration of a continuous diffusion model using a Stochastic Differential Equation (SDE) that transforms a prior noise distribution into the data distribution. The stochastic nature of the sampling process in continuous diffusion models allows samples generated from a single initial point to span the entire space  (green line), but this feature limits its practical applicability. To overcome this limitation, probability flow is introduced, which ensures that the generated sample from an initial point follows a deterministic path (red line).  This enhancement enables the continuous diffusion model to be more manageable and applicable in a broader range of scenarios.

In this paper, our concentration is primarily on discrete diffusion models. An example of such a model, based on SDDM with 15 states, is depicted in Figure \ref{fig:DDPM} (b).  Similar to SDE, it is observed that the sampling process is also susceptible to uncertainty (green line). One potential solution could involve incorporating probability flow into discrete diffusion in a similar manner as in the continuous models. Nonetheless, as previously mentioned in the introduction, this is not a viable option in discrete models due to the lack of a deterministic mapping between the latent space and the data space. Thus, there is a necessity for a redefined probability flow that is tailored to discrete diffusion models, and this forms the core of this paper. This study examines the probability flow of discrete diffusion models through the concept of optimal transport. Initially, we demonstrate that the continuous probability flow coincides with the Monge optimal transport map (Proposition 4). We then leverage this result to develop a similar probability flow for discrete diffusion models using optimal transport (Proposition 6). Finally, we propose a novel sampling methodology for discrete models that significantly reduces the uncertainty (red line) in the sampling process.

\section{Definitions and Theorems Employed in this Appendix} 

For the sake of reader convenience, we hereby provide a comprehensive list of the definitions and theorems utilized in this paper. Additionally, we limit our representation to the case within $\mathbb{R}^n$.

\paragraph{Theorem 1.} (Theorem 1.48 in \cite{santambrogio2015optimal}) \textit{Suppose that $\mu$ is a probability measure on $(\mathbb{R}^n, \mathcal{B})$ such that $\int |x|^2\intd\mu(x) < \infty$ and that $u: \mathbb{R}^n \rightarrow \mathbb{R} \cup \{+\infty\}$ is convex and differentiable $\mu {\text -}a.e$. Set $\text{\normalfont T} = \nabla u$ and suppose $\int |\text{\normalfont T}(x)|^2\intd\mu(x) < \infty$. Then {\normalfont T} is optimal for the transport cost $c(x, y) = \frac{1}{2} |x - y| ^ 2$ between the measures $\mu$ and $\nu = \text{\normalfont T}_{\texttt{\#}}\mu$.}

\paragraph{Definition 2.} \textit{The optimization problem under constraint is formally defined as follows:}
\begin{equation} \label{con_min}
  \min_{x \in \mathbb{R}^n} f(x) ~~~~\text{subject to} \begin{cases}
    c_i(x) = 0, i \in \mathcal{E} \\
    c_i(x) \geq 0, i \in \mathcal{I}.
  \end{cases}
\end{equation}
\textit{The Lagrangian for this constrained optimization problem is defined as:}
\begin{equation}
  \mathcal{L}(x, \lambda) = f(x) - \sum_{i \in \mathcal{E} \cup \mathcal{I} } \lambda_i c_i(x).
\end{equation}
\textit{Here, $\lambda_i$ represents the Lagrange multiplier associated with the $i^{th}$ constraint.  The active set at any feasible $x$ is defined as the union of the set $\mathcal{E}$ with the indices of the active inequality constraints, that is:}
\begin{equation}
  \mathcal{A}(x) = \mathcal{E} \cup \{i \in \mathcal{I} : c_i(x) = 0\}.
\end{equation}

\paragraph{Definition 3.} (Definition 12.1 in \cite{nocedal1999numerical})
\textit{Given the point $x^*$, we say that the Linear Independence Constraint Qualification (LICQ) holds if the set of active constraint gradients $\{ \nabla c_i(x^*), i \in \mathcal{A}(x^*)\}$ is linearly independent.}

\paragraph{Theorem 4.} (Theorem 12.1 in \cite{nocedal1999numerical}, the Karush-Kuhn-Tucker (KKT) conditions) \textit{Suppose that $x^*$ is a local solution of the problem (\ref{con_min}) and that the LICQ holds at $x^*$. Then there is a Lagrange multiplier vector $\lambda^*$ , with components $\lambda_i^*, i \in \mathcal{E} \cup \mathcal{I}$, such that the following conditions are satisfied at $(x^*, \lambda^*)$}
\begin{subequations}
\begin{align}
  \nabla_x \mathcal{L}(x^*, \lambda^*) = 0, & \label{kkta}\\
  c_i(x^*) = 0, & ~~~~~ \forall i \in \mathcal{E}, \label{kktb} \\
  c_i(x^*) \geq 0, & ~~~~~ \forall i \in \mathcal{I}, \label{kktc} \\
  \lambda^* \geq 0, & ~~~~~ \forall i \in \mathcal{I}, \label{kktd} \\
  \lambda_i^* c_i(x^*) = 0, & ~~~~~ \forall i \in \mathcal{E} \cup \mathcal{I} \label{kkte}.
\end{align}
\end{subequations}

\textit{Remark 5.} According to Theorem 4, the Karush-Kuhn-Tucker (KKT) conditions serve as necessary conditions.  In the case of linear programming, these conditions are not only necessary but also sufficient. To demonstrate this, let us consider the standard form of a linear programming problem:
\begin{equation} \label{lin_prog}
  \min c^T x, ~~~~ \text{subject to} ~ Ax = b, x\geq 0.
\end{equation}
We can write the Lagrangian function for \ref{lin_prog} as
\begin{equation}
  \mathcal{L} (x, \pi, s) = c^T x - \pi^T(Ax - b) - s^T x.
\end{equation}
The KKT conditions are
\begin{subequations} \label{lin_prog_kkt}
\begin{align}
  A^T \pi + s = c, & \label{lin_prog_kkta} \\
  Ax = b, & \label{lin_prog_kktb} \\
  x \geq 0, & \label{lin_prog_kktc} \\
  s \geq 0, & \label{lin_prog_kktd} \\
  x^T s = 0. & \label{lin_prog_kkte}
\end{align}
\end{subequations}
Suppose we have a vector triple $(x^*, \pi^*, s^*)$ that satisfies Equation (\ref{lin_prog_kkt}). In such a scenario, we can deduce that:
\begin{equation}
  c^T x^* = (A^T \pi^* + s)^T x^* = (\pi^*)^T A x^* = b^T \pi^*.
\end{equation}
Let us consider another feasible point denoted by $\bar{x}$, which satisfies the conditions $A \bar{x} = b$ and $\bar{x} \geq 0$. we can conclude that:
\begin{equation} \label{lin_prog_ineq}
  c^T \bar{x} = (A^T \pi^* + s^*)^T \bar{x} = b^T \pi^* + \bar{x}^T s^* \geq b^T \pi^* = c^T x^*.
\end{equation}
The inequality (\ref{lin_prog_ineq}) demonstrates that the KKT conditions serve as sufficient conditions.

\paragraph{Theorem 6.} (Theorem 8.5.1 in \cite{oksendal2013stochastic}) \textit{Let $X_t$ be an It\^{o} diffusion given by}
\begin{equation}
  {\rm d} X_t = b(X_t) \intd t + \sigma(X_t)\intd B_t, ~~ b \in \mathbb{R}^n, \sigma \in \mathbb{R}^{n \times m}, X_0 = x,
\end{equation}
\textit{and let $Y_t$ be an It\^{o} process given by}
\begin{equation}
  {\rm d} Y_t = u(t, \omega)\intd t + v(t, \omega)\intd B_t, ~~ u \in \mathbb{R}^n, v \in \mathbb{R}^{n \times m}, Y_0 = x.
\end{equation}
\textit{Assume that}
\begin{equation}
  u(t, \omega) = c(t, \omega) b(Y_t) ~~ and ~~ vv^T(t, \omega) = c(t, \omega)\sigma \sigma^T(Y_t),
\end{equation}
\textit{for a.a. $t, \omega$. Define $\beta_t$ and $\alpha_t$ as:}
\begin{equation}
  \beta_t = \beta(t, \omega) = \int_0^t c(s, \omega)\intd s ~~ and ~~ \alpha_t = \inf\{s : \beta_s > t\}.
\end{equation}
\textit{Then $Y_{\alpha_t}$ coincides in law with $X_t$, denoted by $Y_{\alpha_t} \simeq X_t$}.

\paragraph{Theorem 7.} (Theorem 4.1 of Chapter V, \S 4 in \cite{lang2012fundamentals}, Poincar\'e's lemma). \textit{Let $U$ be an open ball in $\mathbb{R}^n$ and let $\omega$ be a differential form of degree $\geq 1$ on $U$ such that ${\rm d} \omega = 0$. Then there exists a differential form $\phi$ on $U$ such that ${\rm d} \phi = \omega$.}

\textit{Remarks.} The conclusion remains valid when the open ball $U$ is substituted with the entirety of $\mathbb{R}^n$.

\paragraph{Theorem 8.} (Theorem 4 in \cite{blanes2009magnus}) \textit{The solution of the differential equation $Y' = A(t) Y$ with initial condition $Y(0) = Y_0$ can be written as $Y(t) = \exp(\Omega(t)) Y_0$ with $\Omega(t)$ defined by}
\begin{equation}
  \Omega' = \intd \exp_\Omega^{-1}(A(t)), ~~ \Omega(0) = {\displaystyle O}.
\end{equation}
\textit{where}
\begin{equation}
  \intd \exp_\Omega^{-1}(A(t)) = \sum_0^\infty \frac{B_k}{k!} {\rm ad}_\Omega^k(A),
\end{equation}
\textit{and $B_k$ is the Bernoulli numbers. ${\rm ad}_\Omega^k(A)$ is defined through}
\begin{equation}
  {\rm ad}_\Omega(A) = [\Omega, A], ~~ {\rm ad}_\Omega^j(A) = [\Omega, {\rm ad}_\Omega^{j-1}(A)], ~~ {\rm ad}_\Omega^0(A) = A, ~~~~ j \in \mathbb{N},
\end{equation}
\textit{where $[A, B] = AB - BA$ is the Lie-bracket}.

\textit{Remarks.} If $A(s) A(t) = A(t) A(s), \forall s, t \geq 0$, $\Omega(t)$ has the simple form $\Omega(t) = \int_0^t A(s) \intd s$.

\section{Proofs}

\subsection{Proof of Proposition 1}

\emph{Proof.} By It\^{o} formula:
\begin{equation}
\begin{aligned}
  {\rm d} (\phi_t X_t) = &\phi_t \theta_t X_t\intd t + \phi_t\intd X_t \\
  =& \phi_t \theta_t X_t\intd t - \phi_t \theta_t X_t\intd t + \phi_t \sigma_t \intd B_t \\
  =& \phi_t \sigma_t \intd B_t.
\end{aligned}
\end{equation}
By Theorem 6, $\phi_{\alpha_t} X_{\alpha_t} \simeq Y_t$, which means $X_t$ coincides in law with $\phi_t^{-1} Y_{\beta_t}$ \hfill $\square$

\textit{Remarks.}~ Proposition 1 posits that the Ornstein-Uhlenbeck (OU) process is essentially a scaling of Brownian motion with a change in time. Consequently, the VE SDEs, VP SDEs, sub-VP SDEs in \cite{song2020score} , as well as the models presented in \cite{karras2022elucidating}, can be regarded as equivalent.

\subsection{Proof of Proposition 2}

\paragraph{Lemma B.2.1} \textit{Let $H_t$ be the Hessian matrices $\nabla^2_{x_t} \log p_B(x_t, t)$, then $H_s H_t = H_t H_s, \forall s, t \geq 0$.}

\emph{Proof.}
\begin{equation}
\begin{aligned}
  H_t =& \nabla^2_{x_t} \log p_B(x_t, t) \\
  =& \nabla_{x_t} \frac{\sum_i \exp(-\loggaui)(-\diri)}{\sum_j \exp(-\loggauj)} \\
  =& \underbrace{\sum_i \nabla_{x_t} (\frac{\exp(-\loggaui)}{\sum_j \exp(-\loggauj)}) (-\diri)}_A + \underbrace{\frac{\sum_i \exp(-\loggaui)}{\sum_j \exp(-\loggauj)} (-\frac{1}{t}) I}_B.
\end{aligned}
\end{equation}
$B$ is a scalar matrix, then it commutes with any matrix.
\begin{equation}
\begin{aligned}  
  A =& \underbrace{(\sum_j \exp(-\loggauj))^{-2}}_C \sum_i [\exp(-\loggaui) (-\diri) \sum_j \exp(-\loggauj) \\
  &- \exp(-\loggaui) \sum_j \exp(-\loggauj) (-\dirj)] (- \diri)^T \\
  =& C \sum_{i,j} \exp(-\loggaui -\loggauj) (\frac{x_t - x_i}{t}) (\frac{x_t - x_i}{t})^T \\
  &- C \sum_{i,j} \exp(-\loggaui - \loggauj) (\frac{x_t - x_j}{t}) (\frac{x_t - x_i}{t})^T \\
  =& C \sum_{i < j} \exp(-\loggaui - \loggauj) \frac{1}{t^2} [(x_t - x_i) (x_t - x_i) + (x_t - x_j) (x_t - x_j) \\
  &- (x_t - x_j) (x_t - x_i)^T - (x_t - x_i) (x_t - x_j)^T] \\
  =& C \sum_{i < j} \exp(-\loggaui - \loggauj) \frac{1}{t^2} (x_j - x_i) (x_j - x_i)^T.
\end{aligned}
\end{equation}
As ${x_i}$s lie on the same line, $x_j-x_i$ can be denoted by $x_j-x_i = C_{i,j}v, \forall i, j$, where $v$ is a fixed vector. It has $(x_j-x_i)(x_j-x_i)^{T}={C^{2}_{i,j}}v{v}^{T}$. It is clear that ${C^{2}_{i,j}}v{v}^{T}$ and ${C^{2}_{k,l}}v{v}^{T}$ commutes $\forall i, j, k, l$. Furthermore, as $t$ and $x_t$ only appear in the coefficients, $H_t$ and $H_s$ commute with one another. \hfill $\square$

\textit{Proof of Proposition2.} If $Y_0$ follows the initial condition (\ref{init_dist}), and $x_i$s lie on the same line, $Y_t$ will governed by the equation (\ref{conti_prob_flow}). Employing the trick in \cite{chen2018neural}, For a fixed $T$, we define
\begin{equation}
  a(t) = \nabla_{\hat{Y}_t} \hat{Y}_T.
\end{equation}
Then we can derive
\begin{equation}
\begin{aligned}
  \frac{\intd a(t)}{\intd t} = & \lim_{\epsilon \rightarrow 0^+} \frac{a(t + \epsilon) - a(t)}{\epsilon} \\
  = & \lim_{\epsilon \rightarrow 0^+} \frac{a(t + \epsilon) - a(t + \epsilon) \nabla_{\hat{Y}_t} (\hat{Y}_t - \epsilon \frac{1}{2} \nabla_{\hat{Y}_t} \log p_B(\hat{Y}_t, t) + \mathcal{O}(\epsilon^2))}{\epsilon} \\
  = & \lim_{\epsilon \rightarrow 0^+} \frac{\epsilon a(t + \epsilon) \nabla^2_{\hat{Y}_t} \log p_B(\hat{Y}_t, t) + \mathcal{O}(\epsilon^2)}{2 \epsilon}\\
  = & \frac{1}{2} a(t) \nabla^2_{\hat{Y}_t} \log p_B(\hat{Y}_t, t),
\end{aligned}
\end{equation}
where $\nabla^2$ is the \textit{Hessian operator}. Based on Lemma B.2.1, theorem 8 and the fact that $a(T) = \nabla_{\hat{Y}_T} \hat{Y}_T = I$, $a(t) = \nabla_{\hat{Y}_t} \hat{Y}_T$ is symmetric. Then theorem 7 shows that the equation $\nabla_{\hat{Y}_t} u(\hat{Y}_t) = \hat{Y}_T(\hat{Y}_t)$ has a solution. Furthermore, since $a(t)$ is a matrix exponential of a symmetric matrix, it must be positive semi-definite. Consequently, the solution $u$ is convex. According to theorem 1, the map $\hat{Y}_T(\hat{Y}_t)$ is optimal for the quadratic transport cost.  \hfill $\square$

\subsection{Proof of Proposition 3}
\emph{Proof.} The definition of $\tilde{Y}_{t+s}$ is given by Equation (\ref{infi_trans}). It can be observed that the term $\tilde{Y}_t (\tilde{Y}_{t + s})$ on the right-hand side indicates that the evolution speed of $\tilde{Y}_{t+s}$ is constant, which implies that all particles travel at a uniform rate. Consequently, for a given initial condition $\tilde{Y}_t$,
\begin{equation}
  \tilde{Y}_{t + s} = \tilde{Y}_t - \frac{1}{2}\nabla_{\tilde{Y}_t} \log p_B(\tilde{Y}_t) s.
\end{equation}
Further, we have: 
\begin{equation}
  \nabla_{\tilde{Y}_t} \tilde{Y}_{t + s} = I - \frac{1}{2} \nabla^2_{\tilde{Y}_t} \log p_B(\tilde{Y}_t) s.
\end{equation}
It is evident that $\nabla_{\tilde{Y}_t} \tilde{Y}_{t + s}$ is symmetric and for small values of $s$, it is also positive semi-definite. Based on the same reasoning as the proof of Proposition 2, we can conclude that the map $\tilde{Y}_{t+s} (\tilde{Y}_t)$ is optimal for the quadratic transport cost.  \hfill $\square$

\subsection{Proof of Proposition 4}
\emph{proof.}  As Proposition 1 establishes that $X_t = \phi_t^{-1} Y_{\beta_t}$, the single-time marginal distribution $p_{OU}(x_t, t)$ for the Ornstein-Uhlenbeck process can be expressed as follows:
\begin{equation}
  p_{OU}(xt, t) = \frac{1}{N} \sum_i^N (2 \pi \beta_t \phi_t^{-2})^{-\frac{n}{2}} \exp (- \frac{|x_t - \phi_t^{-1} x_i|^2}{2 \beta_t \phi_t^{-2}}).
\end{equation}
The probability flow ODE for Ornstein-Uhlenbeck process is:
\begin{equation}
  \intd \hat{X}_t = [-\theta_t \hat{X}_t - \frac{1}{2} \sigma_t^2 \nabla_{\hat{X}_t} \log p_{OU}(\hat{X}_t, t)] dt.
\end{equation}
We start from $\hat{Y}_t$ with the change of variable $Z_t = \phi_t^{-1} \hat{Y}_{\beta_t}$:
\begin{equation}
\begin{aligned}
  \frac{\intd}{\intd t} Z_t = & \frac{\intd \phi_t^{-1}}{\intd t} \hat{Y}_{\beta_t} + \phi_t^{-1} \frac{\intd \hat{Y}_{\beta_t}}{\intd \beta_t} \frac{\intd \beta_t}{\intd t} \\
  = & - \phi_t^{-1} \theta_t \hat{Y}_{\beta_t} + \phi_t^{-1} (\sigma_t \phi_t)^2 [-\frac{1}{2} \nabla_{\hat{Y}_{\beta_t}} p_B(\hat{Y}_t, \beta_t)] \\
  = & -\theta_t Z_t - \frac{1}{2} \phi_t \sigma_t^2 \frac{\sum_i \exp (-\frac{|\hat{Y}_{\beta_t} - x_i|^2}{2 \beta_t}) \frac{\hat{Y}_{\beta_t} - x_i}{\beta_t}}{\sum_j \exp (-\frac{|\hat{Y}_{\beta_t} - x_j|^2}{2 \beta_t})} \\
  = & -\theta_t Z_t - \frac{1}{2} \sigma_t^2 \frac{\sum_i \exp(- \frac{|\phi_t^{-1} \hat{Y}_{\beta_t} - \phi_t^{-1} x_i|^2}{2 \beta_t \phi_t^{-2}}) \frac{\phi_t^{-1} \hat{Y}_t - \phi_t^{-1} x_i}{\beta_t \phi_t^{-2}}}{\sum_j \exp(- \frac{|\phi_t^{-1} \hat{Y}_{\beta_t} - \phi_t^{-1} x_j|^2}{2 \beta_t \phi_t^{-2}})} \\
  = & -\theta_t Z_t - \frac{1}{2} \sigma_t^2 \frac{\sum_i \exp(- \frac{Z_t - \phi_t^{-1} x_i}{2 \beta_t \phi_t^{-2}}) \frac{Z_t - \phi_t^{-1} x_i }{\beta_t \phi_t^{-2}} }{\sum_j \exp(- \frac{|Z_t - \phi_t^{-1} x_j|^2}{2 \beta_t \phi_t^{-2}})} \\
  = & -\theta_t Z_t - \frac{1}{2} \sigma_t^2 \nabla_{Z_t} \log p_{OU} (Z_t, t).
\end{aligned}
\end{equation}
As $\phi_0 = 1$ and $\beta_0 = 0$, the initial distribution of $X_0$ and $Z_0$ is the same. Consequently, $\hat{X}_t$ and $Z_t$ follow the same ODE with identical initial conditions. Thus, we have $\hat{X}_t = Z_t = \phi_t^{-1} \hat{Y}_{\beta_t}$ and $\nabla_{\hat{X}_t} \hat{X}_{t+s} = \frac{\phi_{t}}{\phi_{t+s}} \nabla_{\hat{Y}_{\beta_t}} \hat{Y}_{\beta_{t+s}} $, which is symmetric and positive semi-definite by Proposition 2. Therefore, we can conclude that the map $\hat{X}_{t+s} (\hat{X}_t)$ is optimal for the quadratic transport cost. \hfill $\square$

\subsection{Proof of Proposition 5}
\emph{Proof.} For the original process (discrete analogue of Brownian motion), the transition rate is:
\begin{equation}
  {{Q_D}^{i}_{j}=}\begin{cases}
    1, & d_D(i,j)=1,\\
  \sum\limits_{j \in N(i)
  }-{Q_D}^{i}_{j}, & i=j,\\
  0, & others,
  \end{cases}
\end{equation}
where $N(i) = \{k: d_D(i, k) = 1 \}$. The Kolmogorov forward equation of this process is written as:
\begin{equation}\label{pd_ode}
\begin{aligned}
  \frac{d{{P_D}_{i}{(t)}}}{dt}=&\sum\limits_{i^{'}} {P_D}_{i{'}}{(t)}{Q_D}^{i^{'}}_{i} \\
  =& {P_D}_{i}{(t)} \times \sum\limits_{i^{'} \in N(i)}-{Q_D}^{i}_{i^{'}} + \sum\limits_{i^{'} \in N(i)} {P_D}_{i^{'}} {(t)} \times 1 \\
  &+ \sum\limits_{i^{'} \in \{k: d_D(i, k) > 1\}} {P_D}_{i^{'}} {(t)} \times 0 \\
  =& \sum\limits_{i^{'} \in N(i)}({P_D}_{i^{'}}{(t)} - {P_D}_{i}{(t)}).
\end{aligned}
\end{equation}

In contrast, the transition rate of our new process is:
\begin{equation}
  {{Q}^{i}_{j}=} \begin{cases}
    \frac{ReLU({P_D}_{i}{(t)}-{P_D}_{j}{(t)})}{{P_D}_{i}{(t)}}, &  d_D(i,j)=1\\
\sum\limits_{d_D(i,j)=1}-{Q}^{i}_{j}, & i=j\\
0, & others.
  \end{cases}
\end{equation}

Our new process can be written as:
\begin{equation} \label{new_ode}
\begin{aligned}
  \frac{d{P_{i}{(t)}}}{dt} =&\sum\limits_{i^{'}}
  P_{i^{'}}{(t)}
  {Q}^{i^{'}}_{i} \\
  =&P_{i}{(t)} \times \sum\limits_{i^{'} \in N(i)}-{Q}^{i}_{i^{'}} \\
  &+ \sum\limits_{i^{'} \in N(i)} P_{i^{'}}{(t)} \times \frac{ReLU({P_D}_{i^{'}}{(t)}-{P_D}_{i}{(t)})}{{P_D}_{i^{'}}{(t)}}\\ &+\sum\limits_{i^{'} \in \{k: d_D(i, k) > 1\}} P_{i^{'}}{(t)} \times 0 \\
  =&-\sum\limits_{i^{'} \in N(i)} P_i{(t)} \frac{ReLU({P_D}_{i}{(t)}-{P_D}_{i^{'}}{(t)})}{{P_D}_i(t)}  \\
  &+ \sum\limits_{i^{'} \in N(i)} P_{i^{'}}(t) \frac{ReLU({P_D}_{i^{'}}{(t)} - {P_D}_{i}{(t)})}{{P_D}_i^{'}(t)}. 
\end{aligned}
\end{equation}
Substitute $P = P_D$ in Equation (\ref{new_ode}), we get the same form in \ref{pd_ode}, which means $P_D$ also solves the Equation (\ref{new_ode}). Thus, $P_{i}(t)={P_D}_{i}(t)$, $\forall t \geq 0,i \in \{1,2,\cdots,S\}^K$, according to Picard-Lindelöf theorem. \hfill $\square$

\subsection{Proof of Proposition 6}
Let $a = P(t)$, $b = P(t+\varepsilon)$. As our generator only allows flux between adjacent states, we define the transport map $\Pi^{*} \in \mathbb{R}^{k \times k}$ as:
\begin{equation} \label{eq:def_pi_star}
  {\Pi^{*}}^i_j = \int_t^{t+\epsilon} P_i(t) Q^i_j(t) \intd t,
\end{equation}
which is the probability transported from state $i$ to state $j$ in the time interval $[t, t + \epsilon]$. As the probability $P(t)$ is continuous with respect to time $t$, we choose the $\epsilon$ such that the sign of all the quantities $P_i(t) - P_j(t)$ for $\{i,j \in \{1,2,\cdots,S\}^K : d_D(i, j) = 1 \}$ do not change. Under this assumption, the flux directions do not change at every state.

We claim that $\Pi^{*}$ solves the optimal transport problem:
\begin{equation} \label{disc_ot_prob}
\begin{aligned}
  \min\limits_{\Pi} \sum\limits_{i,j}\Pi_{j}^{i}  d_D(i,j), \\
  s.t. \begin{cases}
    &\sum\limits_{i}\Pi^{i} = b, \\
    &\sum\limits_{j}\Pi_{j} = a, \\
    &\Pi \geq 0.
  \end{cases}
\end{aligned}
\end{equation}

\emph{Proof.} The Lagrangian function for this optimization problem is:
\begin{equation}
    L(\Pi, \psi, \phi, \lambda)= \sum\limits_{i,j}\Pi^{i}_{j}d_D(i,j)+\psi_i(\Pi^{i}_{j} - a_i)+\phi_j(\Pi^{i}_{j} - b_j)-\lambda^{i}_{j}\Pi^{i}_{j},
\end{equation}
where $\psi_i\text{,~}\phi_j \text{~and~} \lambda^i_j$ are Lagrange multipliers.
According to Remark 5 and the fact that this is a linear programming, $\Pi^{*}$ is optimal if and only if there exists a set of $\phi_i\text{,~}\psi_j \text{~and~} \lambda^i_j$ that satisfy the following equations for $\forall~i,j$:

\begin{subequations}
  \begin{align}
    d_D(i,j)+\psi_i+\phi_j-\lambda_{j}^{i} &= 0 & \label{eq:lag}\\
    \lambda^i_j &\geq 0 & \label{eq:lambda0}\\
    \lambda^{i}_{j} \Pi^{i}_{j} &= 0 & \label{eq:hubu}\\
    \sum\limits_{i}\Pi^{i}_{j} &= b_{j} & \label{eq:pr1} \\
    \sum\limits_{j}\Pi_{j}^{i} &= a_{i}& \label{eq:pr2} \\
    \Pi^{i}_{j} &\geq 0. \label{eq:pipos}& 
  \end{align}
\end{subequations}

Firstly, we consider the $i,j$ pairs where $d_D(i,j)\leq1$. In this case ${\Pi^{*}}_j^i ~\text{may} ~> 0$ (thus $\lambda^{i}_{j}=0$). Besides, the Equation (\ref{eq:def_pi_star}) indicates that ${\Pi^*}^i_i > 0$, thus we have $\lambda^{i}_{i}=0$. Then the Equation (\ref{eq:lag}) comes to:
\begin{equation}
    \psi_i+\phi_i=0.
    \label{eq:neg}
\end{equation}
According to the construction of our generator $Q$, there is no mutual flux, thus we obtain:
\begin{equation}
\Pi^{i_1,\dots,i_{l},\dots,i_K}_{i_1,\dots,i_{l}+1,\dots,i_K} \neq 0 \text{ or } \Pi^{i_1,\dots,i_{l}+1,\dots,i_K}_{i_1,\dots,i_{l},\dots,i_K} \neq 0, ~~ \forall ~ i.
\end{equation}
By substituting this result into complementary slackness condition \ref{eq:hubu}, we have:
\begin{equation}
\lambda^{i_1,\dots,i_{l},\dots,i_K, }_{i_1,\dots,i_{l}+1,\dots,i_K} = 0 \text{ or } \lambda^{i_1,\dots,i_{l}+1,\dots,i_K}_{i_1,\dots,i_{l},\dots,i_K} = 0.
\end{equation}
Since $d_D([i_1,\dots,i_{l},\dots,i_K], [i_1,\dots,i_{l}+1,\dots,i_K]) = 1$, from Equation (\ref{eq:lag}), we can obtain:
\begin{equation}
1+\psi_{i_1,\dots,i_{l},\dots,i_K}+\phi_{i_1,\dots,i_{l}+1,\dots,i_K} = 0 \text{ or } 1+ \psi_{i_1,\dots,i_{l}+1,\dots,i_K}+\phi_{i_1,\dots,i_{l},\dots,i_K} = 0.
\label{eq:hubu2}
\end{equation}
Solving Equations (\ref{eq:neg}) and (\ref{eq:hubu2}) simultaneously, we get:
\begin{equation}
\left\{
\begin{aligned}
&\psi_{i_1,\dots,i_{l}+1,\dots,i_K} = 1 + \psi_{i_1,\dots,i_{l},\dots,i_K} \\
&\phi_{i_1,\dots,i_{l}+1,\dots,i_K} = -1 + \phi_{i_1,\dots,i_{l},\dots,i_K}
\end{aligned}
\right.
\text{ or }
\left\{
\begin{aligned}
&\psi_{i_1,\dots,i_{l}+1,\dots,i_K} = -1 + \psi_{i_1,\dots,i_{l},\dots,i_K} \\
&\phi_{i_1,\dots,i_{l}+1,\dots,i_K} = 1 + \phi_{i_1,\dots,i_{l},\dots,i_K}
\end{aligned}
\right..
\end{equation}
Therefore, given $\psi_{0,\dots, 0}$, $\psi_{i_1,\dots, i_K}$ and $\phi_{i_1,\dots, i_K}$ can be calculated by:
\begin{align}
\psi_{i_1,\dots, i_K} &= \psi_{0,\dots, 0} + m^{i_1,\dots, i_K}_{0, \dots, 0} - n^{i_1,\dots, i_K}_{0, \dots, 0}, \label{eq:solupsi}\\
\phi_{i_1,\dots, i_K} &= -\psi_{i_1,\dots, i_K}, \label{eq:soluphi}
\end{align}
where $m^{i_1,\dots, i_K}_{0, \dots, 0} + n^{i_1,\dots, i_K}_{0, \dots, 0} = d_D(0, i)$, $m^{i_1,\dots, i_K}_{0, \dots, 0} \in \mathbb{N}_0$, and $n^{i_1,\dots, i_K}_{0, \dots, 0} \in \mathbb{N}_0$. $\mathbb{N}_0$ represents the set of all non-negative integers. The quantity $m^{i_1,\dots, i_K}_{0, \dots, 0}$ is the number where $\Pi^{i_1,\dots,i_{l},\dots,i_K}_{i_1,\dots,i_{l}+1,\dots,i_K} \neq 0$ and $n^{i_1,\dots, i_K}_{0, \dots, 0}$ is the number where $\Pi^{i_1,\dots,i_{l}+1,\dots,i_K}_{i_1,\dots,i_{l},\dots,i_K} \neq 0$. Consequently, we find all the Lagrange multipliers for $d_D (i,j) \leq 1$

Then, we consider $i,j$ pairs when $d_D(i,j)>1$ which indicates ${\Pi^*}^i_j = 0$. We use $\psi_i$ and $\phi_j$ in Equation (\ref{eq:solupsi}) and (\ref{eq:soluphi}). To satisfy the KKT condition, we only need to verify that there is $\lambda^{i}_{j}$ satisfies Equation (\ref{eq:lambda0}) and Equation (\ref{eq:lag}). From Equation (\ref{eq:lag}), $\lambda^{i}_{j}$ can be written as:
\begin{equation}
    \lambda^{i}_{j} = d_D(i,j)+\psi_i+\phi_j \label{eq:lambda}
\end{equation}
Let $r_l=\min(i_l, j_l)$, it has:
\begin{align}
    d_D{(i, j)} &= d_D{(i,r)}+d_D{(j,r)} \\
    \psi_{i} &= \psi_{r} + m_{i}^{r}- n_{i}^{r} \\ 
    \phi_{j} &= -\psi_{j} = -\psi_{r} - m_{j}^{r}+ n_{j}^{r} \\
    d_D{(i,r)} &=m_{i}^{r}+n_{i}^{r} \\
    d_D{(j,r)} &=m_{j}^{r}+n_{j}^{r} \\
    m_{i}^{r},n_{i}^{r},m_{j}^{r},n_{j}^{r}&\geq0.
\end{align}
Substitute the above results to Equation (\ref{eq:lambda}), we have:
\begin{align}
    \lambda^{i}_{j} &=2(m_{i}^{r}+n_{j}^{r}) \geq0
\end{align}

As a result, we find all the Lagrange multipliers. Since Equations (\ref{eq:pr1}) and (\ref{eq:pr2}) are naturally satisfied by the construction of $\Pi^*$, we conclude that the KKT conditions are met at $\Pi^*$:

\textcircled{\small{1}} Primal Feasibility: (\ref{eq:pr1}), (\ref{eq:pr2}), (\ref{eq:pipos})

\textcircled{\small{2}} Dual Feasibility: (\ref{eq:lag}), (\ref{eq:lambda0})

\textcircled{\small{3}} Complementary slackness: (\ref{eq:hubu})

According to Remark 5, the KKT conditions indicate $\Pi^*$ is a solution to the optimal transport problem (\ref{disc_ot_prob}). \hfill $\square$

\subsection{Proof of Proposition 7}
This is a special case of Proposition 6, where the generator $Q$ remains constant throughout time.

\begin{algorithm}[t]
  \SetAlgoNoLine
  \DontPrintSemicolon
  $t \leftarrow T$\;
  $i_t^{1:K} \sim P_T(i_T^{1:K})$\;
  \While{ $t > 0$}{
  Compute matrix $P^\theta_D = {[P^\theta_D[l, j]]}_{K \times S}$, where $P^\theta_D[l, j] = {P^\theta_D}_{i_{t-\tau}^{l}=j | i_{t} \backslash i_{t}^l}(t)$, $l=1,\dots,K$, $j = 1,\dots,S$ with softmax operation on the results of $K \times S$ forward pass of the model\;
  
  Encoded the three candidate states (i.e., $i_{t}^l$, $i_{t}^l - 1$ and $i_{t}^l + 1$) for $i_{t-\tau}^l$ with one-hot code: $O_{stay} \leftarrow I_{K \times K}[i_t]$; $O_{sub} \leftarrow I_{K \times K}[i_t - 1]$; $O_{add} \leftarrow I_{K \times K}[i_t + 1]$\;
  
  Fetch the probability $P^\theta_D$ for the above candidate state: $P_{stay} \leftarrow \sum_j (O_{stay} \circ P^\theta_D)$; $P_{sub} \leftarrow \sum_j (O_{sub} \circ P^\theta_D)$; $P_{add} \leftarrow \sum_j (O_{add} \circ P^\theta_D)$\;
  
  $R_{i_{t-\tau}}^{i_{t}}(t) \leftarrow O_{sub} \circ ReLU(P_{sub}/ P_{stay} - 1) + O_{add} \circ ReLU(P_{add}/ P_{stay} - 1) - O_{stay} \circ (ReLU(P_{sub}/ P_{stay} - 1) + ReLU(P_{add}/ P_{stay} - 1))$ with Equation (\ref{rev_rate})\;
  
  $P^{\theta}(i_{t-\tau}^{l}|i_{t}) \leftarrow \tau R_{i_{t-\tau}}^{i_{t}}(t) + O_{stay}$ with Equation (\ref{euler})\;
  
  $i_{t-\tau} \leftarrow$ Categorical$\Big(P^{\theta}(i_{t-\tau}^{l}|i_{t})\Big)$\;
  
  $t \leftarrow t - \tau$\;
  }
  \caption{Generative Reverse Process with Discrete Probability Flow (DPF) }
  \label{alg:DPF}
\end{algorithm}

\section{Experiment} 
\subsection{Algorithm}
Our training process follows the same procedure as SDDM, with the distinction that our forward process incorporates the rate we formulated in Equation (\ref{ot_Q}) to align with optimal transport theory. The loss function employed during the training process is as follows:
\begin{equation}
\theta^{*} = \arg\min\limits_{\theta}\int_0^{T}\sum\limits_{i_t \in \{1,2,\cdots,S\}^K}q_t(i_t)\left [ \sum\limits_{l=1}^{K}-\log P_t(i_t^{l}|i_t \backslash i_t^{l}) \right] dt.
\end{equation}
The sampling process with the proposed discrete probability flow is shown in Algorithm \ref{alg:DPF}. In our algorithm, as $R$ is non-zero only when $d_D{(i_t,i_{t-\tau})} \leq 1$, the calculation of the reverse transition rate $R$ (as defined in Equation \ref{rev_rate}) is divided into three cases: staying in the current state ( $i^{l}_{t-\tau} = i^{l}_{t}$, i.e., "stay"), jumping to the next state ($i^{l}_{t-\tau} = i^{l}_{t} + 1$, i.e., "add"), and jumping to the previous state ($i^{l}_{t-\tau} = i^{l}_{t} - 1$, i.e., "sub"). By combining the rate in these situations, we can derive $P_{\theta}(i_{t-\tau}^{l}|i_{t})$ from (\ref{euler}) and (\ref{rev_sample}), which allows us to sample the next state accordingly. This process  continues iteratively until $t = 0$.
\begin{table}[!t]
\centering
\caption{Average MMD between different distributions of data.}
\begin{tabular}{c|ccc}
\Xhline{1pt}
State size  & 2        & 5         & 10        \\ \hline
Average MMD & 5.336e-3 & 2.201e-2 & 6.531e-3 \\ \Xhline{1pt}
\end{tabular}
\label{tab:ammd}
\end{table}
\begin{figure*}[!t]
  \centering
  \includegraphics[width=0.85\textwidth]{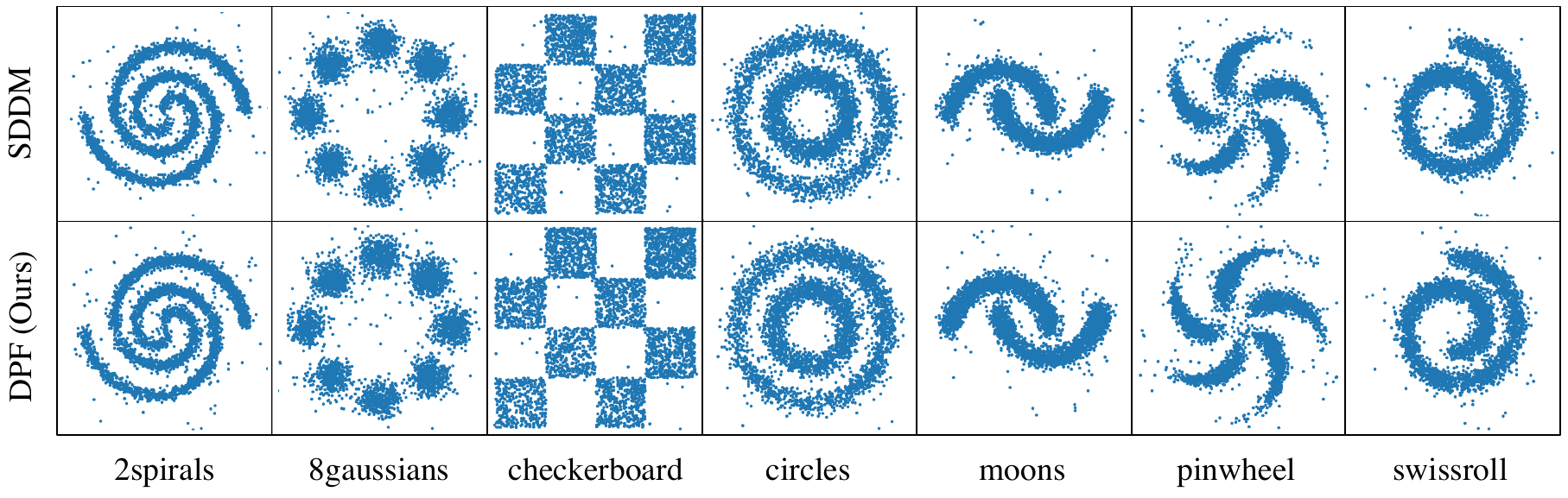} 
  \caption{Visualization of the generation quality on generated samples with state size = 5 for SDDM and DPF. }
  \label{fig:base5_generate}
\end{figure*}
\begin{figure*}[!t]
  \centering
  \includegraphics[width=0.85\textwidth]{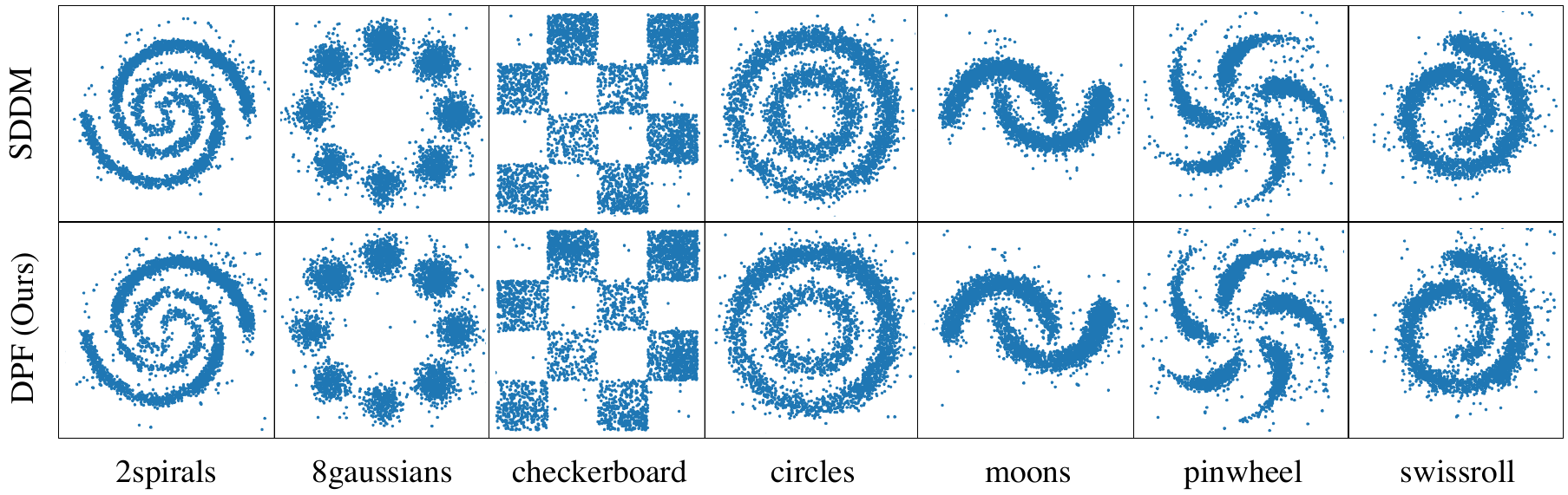} 
  \caption{Visualization of the generation quality on generated samples with state size = 10 for SDDM and DPF. }
  \label{fig:base10_generate}
\end{figure*}

\subsection{Synthetic Dataset}
Following \cite{pmlr-v162-zhang22v,NEURIPS2020_7612936d,sun2022score},  we utilize synthetic data for model validation. Initially, we generate 2D floating-point data from seven distinct distributions using an infinite data oracle. By employing the same settings as \cite{sun2022score}, we convert each dimension of the data into 16-bit Gray code, resulting in a dataset with discrete dimension = 32 and state size = 2. However, it is not sufficient to validate our method solely on the dataset with state size = 2, since $Q$ in Equation (\ref{ot_Q}) does not cover cases where $d_D(i,j)>1$. Therefore, we further transform the same data into 8-bit 5-base code and 6-bit decimal code respectively, thereby creating two additional datasets: one with dimension = 16 and state size = 5, and another with dimension = 12 and state size = 10.

\subsection{Experiment Details}
In the experiments, our neural network consists of a 3-layer MLP with 256 channels \cite{pmlr-v162-zhang22v, sun2022score}.  We employ the Adam optimizer with a learning rate of 1e-4. The model is trained on a single NVIDIA Quadro RTX 8000, utilizing a batch size of 128 for 300,000 iterations. During training, the parameter $t$ is uniformly sampled from the range of 0 to 1. For the sampling process, the data is generated through 1,000 steps (i.e.,$\tau$ is set to 0.001).

\begin{figure*}[!t]
  \centering
  \includegraphics[width=0.85\textwidth]{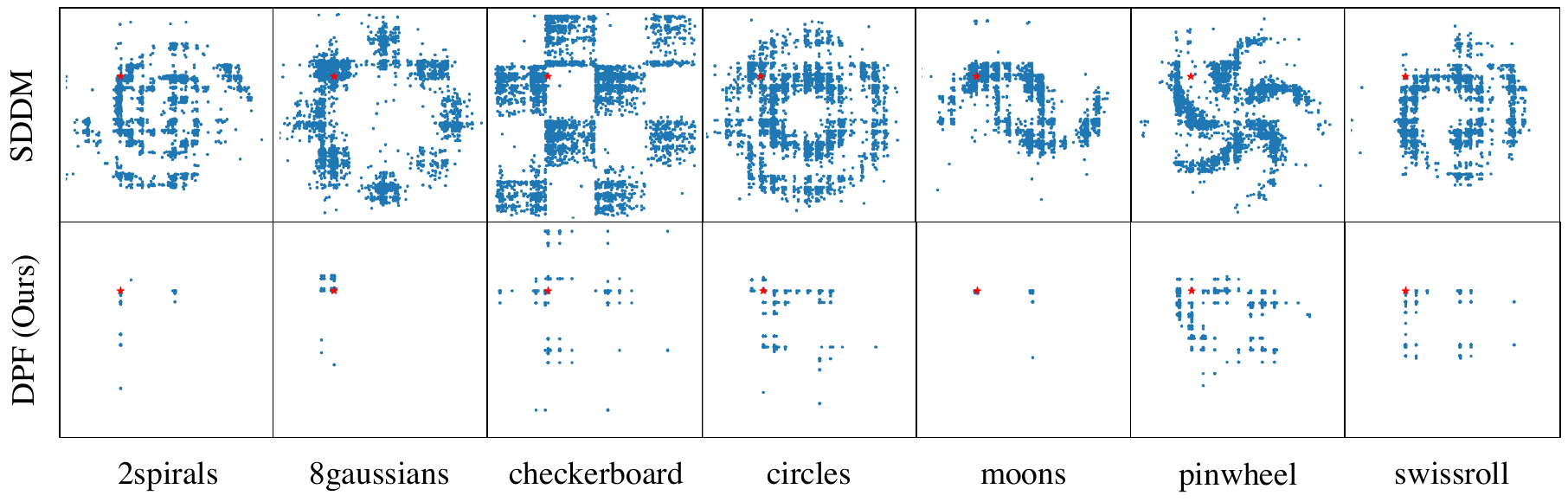} 
  \caption{Visualization of the generating certainty on generated samples with state size = 5 for SDDM and DPF. All the samples (in blue) are randomly generated from the single initial point (in red).}
  \label{fig:base5_variance}
\end{figure*}
\begin{figure*}[!t]
  \centering
  \includegraphics[width=0.85\textwidth]{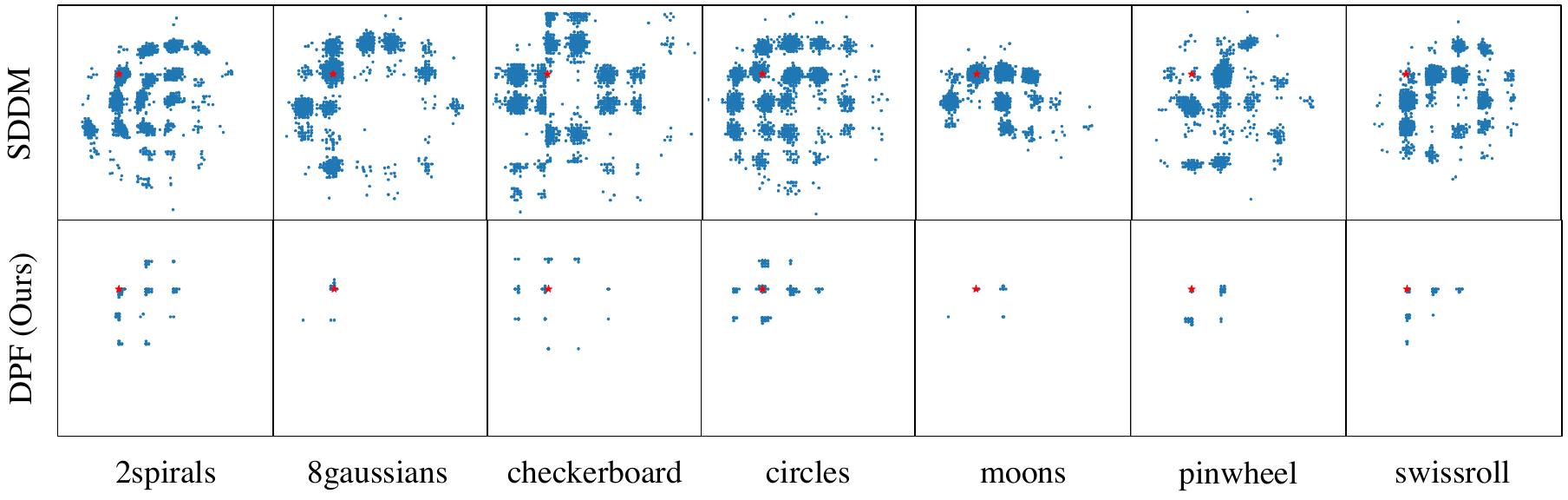} 
  \caption{Visualization of the generating certainty on generated samples with state size = 10 for SDDM and DPF. All the samples (in blue) are randomly generated from the single initial point (in red).}
  \label{fig:base10_variance}
\end{figure*}

\subsection{Quality of Generated Samples}
To evaluate the sampling quality, we generate 40,000 samples for binary data, and 4,000 samples for other type of synthetic data by SDDM and our method. Then we compare these generated samples to true data using Laplace MMD. This evaluation is repeated 10 times, and the average results are presented in Table \ref{tab:mmd}. It is worth noting that the unbiased estimation of MMD \cite{gretton2012kernel} is an approximation by Monte Carlo method, which may cause negative results. It is observed that MMD score of our method is slightly higher than that of SDDM. This is mainly caused by the approximation of $Q_t$. In the sampling process, two terms are present on the right-hand side of Eq. \ref{orig_dis_rev}: $\frac{q_t(y)}{q_t(x)}$ and $Q_t(y, x)$. In SDDM, only one term, i.e., $\frac{q_t(y)}{q_t(x)}$ is estimated using a neural network, as ${Q_D}_t(y, x)$ is known. Different from SDDM, both terms in our method are evaluated using quantities approximated by the neural network, since our $Q_t(y, x)$ is dependent on $\frac{q_t(y)}{q_t(x)}$ (Eq. \ref{eq:27}). This approximation may lead to slightly inferior quality than the SDDM using precise ${Q_D}_t(y, x)$. Due to this being a neural network fitting error, we currently have no feasible alternative approximations to achieve a superior outcome.

To assess the significance of these differences, we presented the MMD between different distributions of real data in Table \ref{tab:ammd}. Taking this result as a reference, we can find that the gap of the MMD score between DPF and SDDM is very small, which is not enough to affect the quality of the generation. This conclusion is further supported by the visualization of the generated samples in Figure \ref{fig:fig1}, Figure \ref{fig:base5_generate} and Figure \ref{fig:base10_generate} also confirm this point, which demonstrates that DPF produces samples of comparable quality to SDDM.

\setlength{\tabcolsep}{1.5mm}{
\centering
\begin{table}[!t]
\centering
  \caption{Comparison of the average $L_1$ distance between the generated samples and initial point. Lower values indicate that the generated sample is closer to initial point.}
  \renewcommand\arraystretch{0.8}
  \begin{tabular}{cccccccc}
  \Xhline{1pt}
  \multicolumn{1}{c|}{}     & 2spirals & 8gaussians & checkerboard & circles & moons & pinwheel & swissroll \\ \Xhline{1pt}
                            & \multicolumn{7}{c}{discrete dimension = 32, state size = 2}              \\ \hline
  \multicolumn{1}{c|}{SDDM} & 13.5595 & 13.3025  & 13.4710 & 13.5848 & 13.6485  & 13.4875 & 13.6962   \\
  \multicolumn{1}{c|}{DPF (ours)}   & 1.5965 & 1.3855  & 0.7525 & 1.1875 & 1.8693  & 1.8135 & 1.6955    \\ \hline
                            & \multicolumn{7}{c}{discrete dimension = 16, state size = 5}              \\ \hline
  \multicolumn{1}{c|}{SDDM} & 12.7220         & 12.4698           & 12.4833  & 12.5390             & 12.6745        & 12.6238      & 12.7510         \\
  \multicolumn{1}{c|}{DPF (ours)}   &  1.5265        &  1.6155          & 0.7090       & 1.1668      & 1.7038        & 1.8088      & 1.5888         \\ \hline
                            & \multicolumn{7}{c}{discrete dimension = 12, 
  state size = 10}             \\ \hline
  \multicolumn{1}{c|}{SDDM} & 11.3433         & 11.0083           & 11.0243  & 11.2205             & 11.6850        & 11.3895      & 11.6333         \\
  \multicolumn{1}{c|}{DPF (ours)}   &  1.7655        &  1.1940          & 0.7143             & 1.1588        & 2.0493      & 1.9283         & 1.7695          \\ \Xhline{1pt}
  \end{tabular}
  \label{tab:dis}
\end{table}
} 
\begin{figure*}[!t]
  \centering
  \includegraphics[width=1\textwidth]{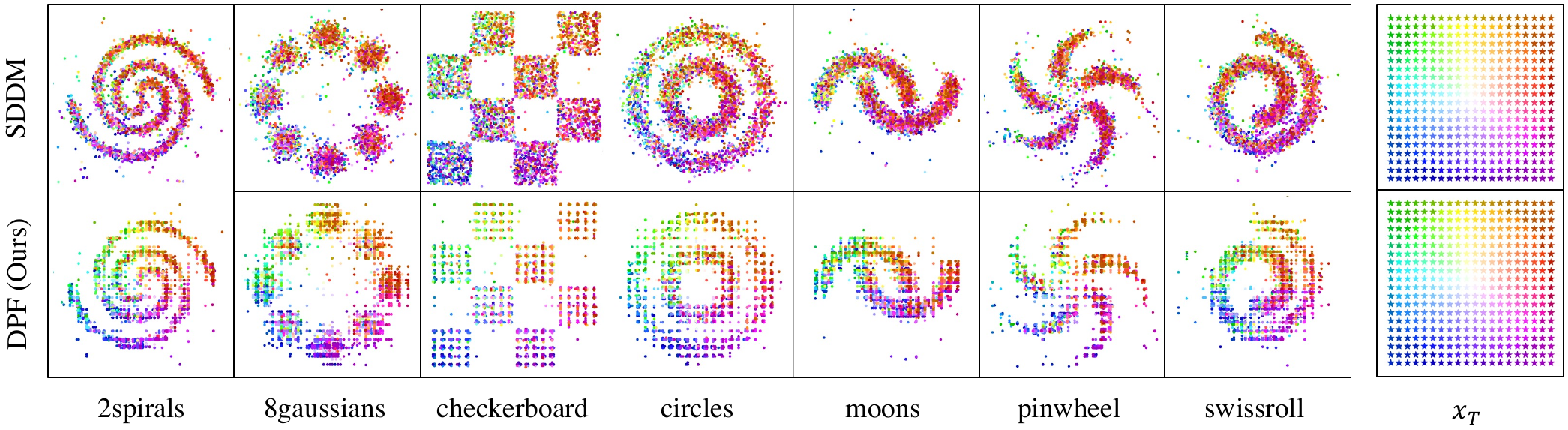} 
  \caption{Visualization of the generated samples with state size = 5 from the given initial points $\bm{x}_T$. 
  Different colors distinguish the generated samples from different initial points $\bm{x}_T$.}
  \label{fig:base5_color}
\end{figure*}
\begin{figure*}[!t]
  \centering
  \includegraphics[width=1\textwidth]{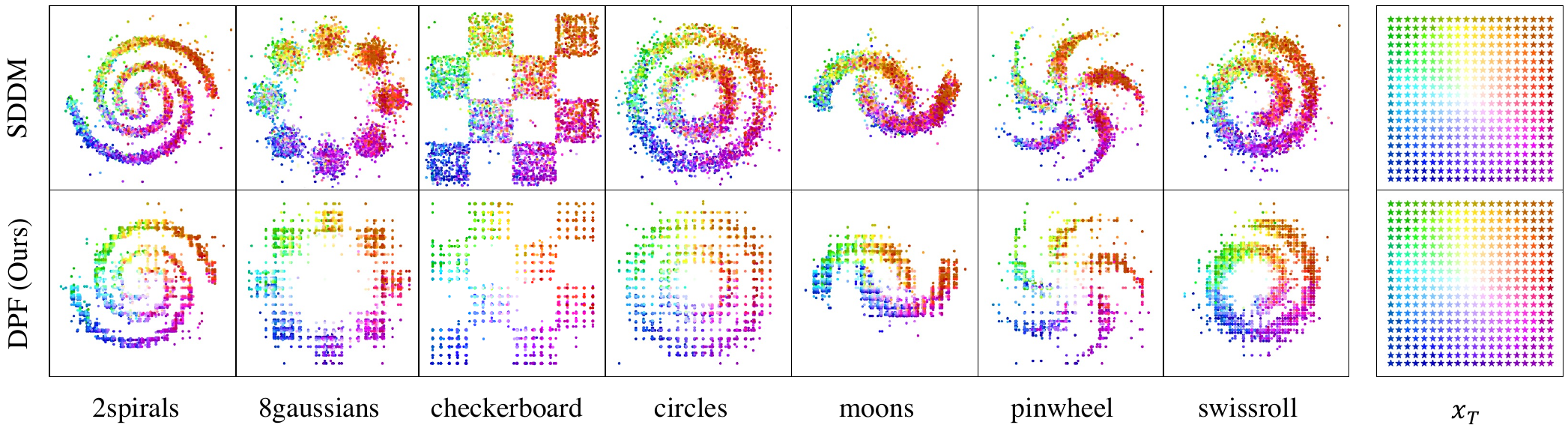} 
  \caption{Visualization of the generated samples with state size = 10 from the given initial points $\bm{x}_T$. 
  Different colors distinguish the generated samples from different initial points $\bm{x}_T$.}
  \label{fig:base10_color}
\end{figure*}

\subsection{Standard Deviation of Generated Samples}
To evaluate the certainty of generated samples, we randomly select a 2D float-point and fix it as the initial point. In this experiments, we use the same initial point (-1.91, 1.57). In the binary case, the point is converted into Gray code, whereas in the 5-base and decimal cases, the original code is utilized. For each dataset, we generate 4,000 points and compute the Expectation of the Conditional Standard Deviation (\ref{eq:var}). Our DPF method results in a significant reduction in the $CSD$ score, as presented in Table \ref{tab:ecv}. For example, on the checkerboard dataset with $S = 5$, our DPF achieves the best score of 1.4103, which is 89\% lower than the score achieved by SDDM. We also visualize these results in Figure \ref{fig:fig2}, Figure \ref{fig:base5_variance}, and Figure \ref{fig:base10_variance}, where the red star represents the initial point, and the blue points denote the generated samples. Furthermore, it is evident that SDDM generates samples from a single initial point across the entire space, especially for datasets with $S=2$. In contrast, our method can only reach a limited number of states from the initial point, indicating the superior sampling certainty of our approach.

We can also observe that our sampling results tend to form rectangles in Figure \ref{fig:fig2}, Figure \ref{fig:base5_variance}, and Figure \ref{fig:base10_variance}. This phenomenon arises from the construction of our synthetic dataset. Specifically, we construct the synthetic dataset states and dimensions by encoding the $x$-axis and y-axis coordinates of the toy dataset (normalized to [0, 1]) into $K/2$-bit $S$-ary. This is equivalent to dividing the data space into rectangle regions, where the first few dimensions determine the approximate location of the data. Since our proposed method significantly reduces the uncertainty, each dimension (including the first few dimensions) has only a limited number of possible values. As a result, the points in Figure \ref{fig:fig2}, Figure \ref{fig:base5_variance}, and Figure \ref{fig:base10_variance} appear to form rectangles.

\subsection{Generated Samples from Different Initial Points}
To display the generated samples from various initial points, we select a $20 \times 20$ grid of initial points and mark them with distinct colors. Subsequently, we generate 10 samples for each initial point and presented the results in Figure \ref{fig:fig3}, Figure \ref{fig:base5_color}, and Figure \ref{fig:base10_color}. It is apparent that the samples obtained through SDDM sampling are mixed together. In contrast, the results obtained by our method exhibit strong regularity, with the generated samples clustering together based on their respective colors. This observation suggests that our method offers improved certainty in the sampling process.

\setlength{\tabcolsep}{1.5mm}{
\centering
\begin{table}[!t]
\centering
  \caption{Comparison of average trajectory length. Lower value indicates a better transport plan.}
  \renewcommand\arraystretch{0.8}
  \begin{tabular}{cccccccc}
  \Xhline{1pt}
  \multicolumn{1}{c|}{}     & 2spirals & 8gaussians & checkerboard & circles & moons & pinwheel & swissroll \\ \Xhline{1pt}
                            & \multicolumn{7}{c}{discrete dimension = 32, state size = 2}              \\ \hline
  \multicolumn{1}{c|}{SDDM} & 32.0075 & 31.7275  & 31.8010 & 32.0258 & 32.0240  & 31.8650 & 31.9988   \\
  \multicolumn{1}{c|}{DPF (ours)}   & 1.6135 & 1.3980  & 0.7640 & 1.1995 & 1.8883  & 1.8265 & 1.7065    \\ \hline
                            & \multicolumn{7}{c}{discrete dimension = 16, state size = 5}              \\ \hline
  \multicolumn{1}{c|}{SDDM} & 26.0680         & 25.3258           & 25.9708  & 25.7565             & 25.8815        & 25.9793      & 26.0810         \\
  \multicolumn{1}{c|}{DPF (ours)}   &  1.5425        &  1.6390          & 0.7275       & 1.1758      & 1.7102        & 1.8178      & 1.5973         \\ \hline
                            & \multicolumn{7}{c}{discrete dimension = 12, 
  state size = 10}             \\ \hline
  \multicolumn{1}{c|}{SDDM} & 21.8558         & 21.7828           & 21.7778  & 21.9100             & 22.3180        & 21.9985      & 22.2688         \\
  \multicolumn{1}{c|}{DPF (ours)}   &  1.7835        &  1.1995          & 0.7213             & 1.1793        & 2.0623      & 1.9433         & 1.7870          \\ \Xhline{1pt}
  \end{tabular}
  \label{tab:tra}
\end{table}
} 

\subsection{Distance Between the Generated Samples and Initial Points}
Our DPF is designed based on the theory of optimal transport, as demonstrated in Proposition 6. Here, we aim to reflect this finding through experimentation as well. To accomplish this, we utilize the generated samples from  Figure \ref{fig:fig3}, Figure \ref{fig:base5_color} and Figure \ref{fig:base10_color}, and calculate the average $L_1$ distance from the generated samples to the corresponding initial point:
\begin{equation}
    d_{D}(i(0),i(T)) = \sum\limits_{l=1}^{S} |i^l(0)-i^l(T)|.
\end{equation}
The results are presented in Table \ref{tab:dis}. It is evidence that DPF greatly reduces the distance between the generated samples and the initial point. Moreover, combined with the visualization results in Figure \ref{fig:fig3}, Figure \ref{fig:base5_color} and Figure \ref{fig:base10_color},  we observe that our method's sampling outcomes tend to concentrate around the high probability states near the initial point. This outcome aligns with our optimal transport design, further verifying the efficacy of our approach. 

However, there is an illusion that the difference between SDDM and PDF decreases as the state size increases. This is mainly because that the Figure \ref{fig:fig3}, Figure \ref{fig:base5_color} and Figure \ref{fig:base10_color} are visualized in the 'float space' instead of the 'encoding space'.  Specifically, our synthetic data with a state size of and a dimension size of is established by encoding the $x$ and $y$ coordinates of the toy dataset (normalized to [0, 1]) to $K/2$-bit $S$-ary respectively. In this encoding, the first dimension of the encoding has the greatest impact on the data position. For example, in binary encoding (state size = 2), the first bit divides the data space into two parts, and determines the part in which it resides. However, as the number of states increases, the space is divided into more parts, and the small change of the first bit can not significantly change the position of the number it represents. This will lead to a narrowing of the gap between our DPF and SDDM in the visualization. Therefore, in such situations, the quantitative results in Table \ref{tab:ecv} are more appropriate for verifying the reduction of uncertainty in the encoding space.

\setlength{\tabcolsep}{1.5mm}{
\centering
\begin{table}[!t]
\centering
  \caption{Comparison of transport efficiency. Larger values indicate better transport efficiency.}
  \renewcommand\arraystretch{0.8}
  \begin{tabular}{cccccccc}
  \Xhline{1pt}
  \multicolumn{1}{c|}{}     & 2spirals & 8gaussians & checkerboard & circles & moons & pinwheel & swissroll \\ \Xhline{1pt}
                            & \multicolumn{7}{c}{discrete dimension = 32, state size = 2}              \\ \hline
  \multicolumn{1}{c|}{SDDM} & 42.36\% & 41.93\%  & 42.36\% & 42.42\% & 42.62\%  & 42.33\% & 42.80\%   \\
  \multicolumn{1}{c|}{DPF (ours)}   & 98.95\% & 99.11\%  & 98.49\% & 99.00\% & 98.99\%  & 99.29\% & 99.36\%    \\ \hline
                            & \multicolumn{7}{c}{discrete dimension = 16, state size = 5}              \\ \hline
  \multicolumn{1}{c|}{SDDM} & 48.80\%         & 49.24\%           & 48.07\%  & 48.68\%             & 48.97\%        & 48.59\%      & 48.89\%         \\
  \multicolumn{1}{c|}{DPF (ours)}   &  98.96\% & 98.56\% & 97.46\% & 99.23\% & 99.63\% & 99.50\% & 99.47\%         \\ \hline
                            & \multicolumn{7}{c}{discrete dimension = 12, 
  state size = 10}             \\ \hline
  \multicolumn{1}{c|}{SDDM} & 51.90\%         & 50.54\%          & 50.62\%  & 51.21\%             & 52.36\%        & 51.77\%      & 52.24\%         \\
  \multicolumn{1}{c|}{DPF (ours)}   &  98.99\% & 99.54\% & 99.02\% & 98.26\% & 99.36\% & 99.22\% & 99.02\%          \\ \Xhline{1pt}
  \end{tabular}
  \label{tab:eff}
\end{table}
}

\begin{figure*}[!t]
  \centering
  \includegraphics[width=1\textwidth]{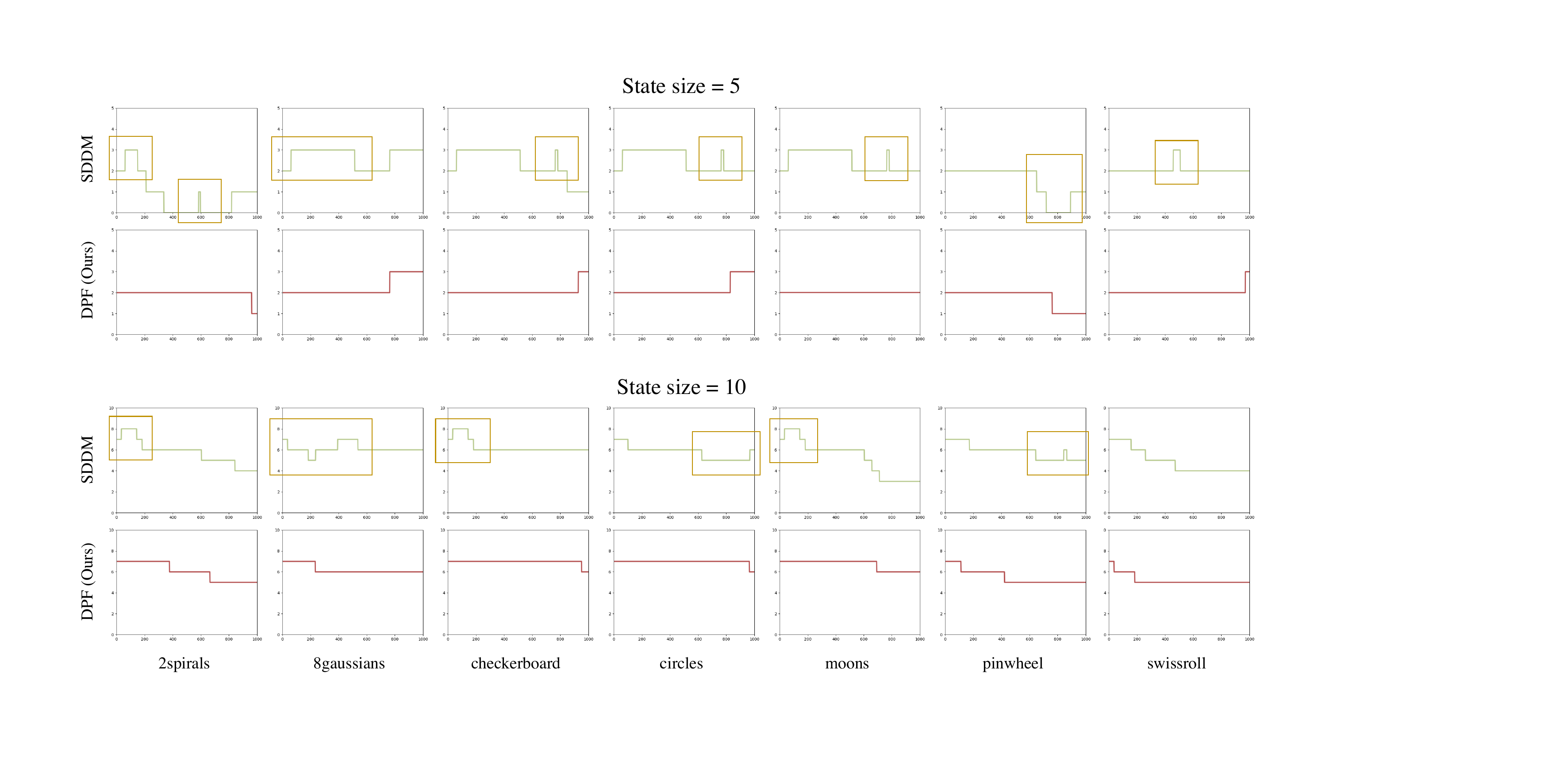} 
  \caption{Visualization of the sampling trajectory. The yellow box highlights the duplicated trajectories encountered during the sampling process.}
  \label{fig:vis_tra}
\end{figure*}

\subsection{Sampling Trajectory Length}
Merely examining the distance between the initial point and the generated samples is insufficient to verify that our DPF aligns with optimal transport principles, as the sampling process may follow different trajectories. Therefore, we calculate the cumulative consumption of the sampling trajectory in Figure \ref{fig:fig3}, Figure \ref{fig:base5_color} and Figure \ref{fig:base10_color} according to the following formula:
\begin{equation}
    d_{tra}(i(0),\dots,i(T)) = \sum\limits_{t \in \{\tau,2\tau \dots, T\}}d_D(i(t), i(t-\tau)).
\end{equation}
The results are shown in Table \ref{tab:tra}. It is evidence that there is a significant decrease in the trajectory length of DPF compared to the SDDM. For instance, our DPF achieves the best score on the checkerboard dataset with S = 5 with a score of 0.7640, which is 97\% lower than the score of SDDM. This suggests that the consumption during our sampling process is lower, which is in line with the optimal transport design.

\subsection{Transport Efficiency}
We also examine the transport efficiency during the sampling process, which can be calculated as the ratio of $L_1$ distance between the initial point and generated sample to the sampling trajectory length:
\begin{equation}
    E_{(i(0),\dots,i(T))} = \frac{d_D(i(0),i(T))}{d_{tra}(i(0),\dots,i(T))}.
\end{equation}
A higher value indicates a more optimal sampling trajectory selected by the model from the initial point to the generated sample, i.e., the higher transport efficiency. The results are presented in Table \ref{tab:eff}. Notably, we observed that only approximately 50\% of the trajectory length of SDDM contributes to the actual distance between the initial point and generated samples. In contrast, the transport efficiency of our DPF is close to 100\%, which means most jumps in our trajectory efficiently contribute to the final transition. This finding demonstrates that the transport plan selected by our DPF is more effective, aligning with our theoretical derivation.

\subsection{Visualization of Sampling  Trajectory}
The visualization of the sampling trajectory for the 0-th dimension of the dataset is shown in Figure \ref{fig:vis_tra}. It is evident that the sampling trajectory of SDDM often exhibits duplicate trajectories, which is also the reason for the low transport efficiency of SDDM in Table \ref{tab:eff}. In contrast, our method, which adheres to the principles of optimal transport theory, ensures that the sampling process only moves toward high probability states, thereby avoiding the occurrence of duplicate trajectories.
 
\subsection{Higher dimension or state scenarios}
To further verify our method is still applicable in higher dimension or state scenarios, we increased the number of states  and dimension to 50 for experiments. Specifically, we set $S = 50, K = 20$ and $S = 5, K = 50$ to avoid dimension redundancy in the $K/2$-bit $S$-ary encoding for the toy dataset (float64) coordinates (i.e., $50^{20/2}<2^{64}$ and $5^{50/2}<2^{64}$). The results of this experiment are shown in Table \ref{tab:high}, which clearly demonstrate that our method can significantly reduce sampling uncertainty even with larger state and dimension sizes.

\subsection{Image Modeling}
In addition to the transition rate designed in Eq. \ref{chow_Q}, our method can also be extended to a broad range of transition rates. For example, we can extend our discrete probability flow to the method in \cite{campbell2022continuous}.
For general $Q_{D_t}$ with ${Q_D}^i_j(t)={Q_D}^j_i(t)$, define:
\begin{equation}
Q^i_j(t)= \begin{cases}
    {Q_D}^i_j(t)\frac{{ReLU}(P_{D_i}(t) - P_{D_j}(t))}{P_{D_i}(t)}, &i \neq j, \\
    -\sum_{j \neq i} Q^i_j, & i = j.
\end{cases}
\end{equation}

\setlength{\tabcolsep}{1.8mm}{\begin{table}[t!]
\centering
  \caption{Application on higher dimension or state scenarios. Lower $CSD$ indicate superior certainty.}
  \renewcommand\arraystretch{0.8}
  \begin{tabular}{cccccccc}
  \Xhline{1pt}
  \multicolumn{1}{c|}{}     & 2spirals & 8gaussians & checkerboard & circles & moons & pinwheel & swissroll \\ \Xhline{1pt}
                            & \multicolumn{7}{c}{discrete dimension = 20, state size = 50}              \\ \hline
  \multicolumn{1}{c|}{SDDM} & 25.8777	& 26.5288	& 25.5106	& 25.7398	& 25.6984	& 25.6984	& 6.4767   \\
  \multicolumn{1}{c|}{DPF (ours)}   & 2.7113 & 4.4274 & 2.6217 & 3.7554 & 3.0774 & 3.3054 & 3.8183    \\ \hline
                            & \multicolumn{7}{c}{discrete dimension = 50, state size = 5}              \\ \hline
  \multicolumn{1}{c|}{SDDM} & 47.2706 & 47.5810 & 47.4964 & 47.2733 & 47.0047 & 46.9103 & 46.9819         \\
  \multicolumn{1}{c|}{DPF (ours)}   &  2.0335 & 1.8134 & 0.7418 & 1.7143 & 1.2840 & 1.4245 & 1.5720
         \\ \Xhline{1pt}
  \end{tabular}
  \label{tab:high}
\end{table}
}

${Q_{D_t}}$ and ${Q_t}$ have the same single-time marginal distribution. Let $q_t = P_D(t)$ and $x \neq y$, the reverse transition rate can be written as:
\begin{subequations}
\begin{align}
R_t(x,y) &= \frac{q_t(y)}{q_t(x)}Q_t(y,x) \\
&= \frac{q_t(y)}{q_t(x)}Q_{D_t}(y,x)\frac{ReLU(q_t(y)-q_t(x))}{q_t(y)} \\
&=Q_{D_t}(y,x)RELU(\frac{q_t(y)}{q_t(x)} - 1)
\end{align}
\end{subequations}

In the same way, the reverse rate in the paper \cite{campbell2022continuous} can be written into the following form:
\begin{equation}
\hat{R}_t^{1:D}(\bm{x}^{1:D}, \tilde{\bm{x}}^{1:D}) = \sum_{d=1}^D R_t^d(\tilde{x}^d, x^d) \delta_{\bm{x}^{1:D \backslash d}, \tilde{\bm{x}}^{1:D \backslash d}} RELU(\sum_{x_0^d} q_{0|t}(x_0^d | \bm{x}^{1:D}) \frac{q_{t|0}(\tilde{x}^d | x_0^d)}{q_{t|0}(x^d | x_0^d)}-1)
\end{equation}

\begin{table}[!t]
\centering
\caption{Comparison of certainty for 
$\tau$LDR-0 and DPF on the Cifar-10 dataset. Here, $CSD$
, class-std, and class-entropy are calculated on 1,000 initial points, each of which has 10 generated images. Lower values indicate superior certainty.}
\begin{tabular}{c|ccc}
\Xhline{1pt}
  & $CSD$        & class-std	& class-entropy        \\ \hline
$\tau$LDR-0 \cite{campbell2022continuous} & 57.6898 & 2.6628 & 1.7703 \\
DPF (ours) & 9.4420 & 1.1819 & 0.5291 \\\Xhline{1pt}
\end{tabular}
\label{tab:cifar}
\end{table}

In this way, the mutual flow between states is eliminated, greatly reducing the sampling uncertainty. To validate this, we validated our DPF on the CIFAR-10 dataset, using the pre-trained discrete diffusion model provided by the paper \cite{campbell2022continuous}. Firstly, we selected 1,000 initial points, and sampled 10 images from each initial point. To measure the sampling certainty on the image data, we used a pre-trained CIFAR-10 classifier to classify the image, and introduce two new metrics, i.e., class-std and class-entropy. The class-std calculates the standard deviation of the categories of the images sampled from the same initial point. While the class-entropy calculates the entropy of the category distribution of the images sampled from the same initial point. Lower class-std and class-entropy indicate better sampling certainty. The experimental results, shown in Table \ref{tab:cifar}, demonstrate that our method can significantly reduce the sample uncertainty compared to the $\tau$LDR-0 method. Additionally, we visualized the sampled images in Fig. \ref{fig:cifar}. It was clear that from an initial point, our method samples almost the same images, while the original sampling method obtains totally different images. 

\section{Discussion}

\subsection{Narrow time interval limited in Proposition 6.}
We limit the time frame to a narrow interval, as the validity of the proof hinges on the constancy of the sign of $P_i(t) - P_j(t)$.
Alternatively, if both equations in Eq. (70) are established concurrently, a contradiction arises whereby 2 equals 0.
Consequently, the KKT condition cannot be satisfied by any suitable Lagrange multipliers, thereby rendering the plan sub-optimal.

From an intuitive standpoint, DPF only avoids instantaneous mutual flow, which does not ensure the elimination of mutual flow during finite interval. For example, if we assume $P_i(t) > P_j(t)$ in the interval $[t,t + \epsilon / 2)$ and $P_i(t) < P_j(t)$ in $(t+\epsilon/2, t+\epsilon]$, it follows that $\Pi^i_j > 0$ and $\Pi^j_i > 0$. Assuming $\Pi^i_j > \Pi^j_i$, we can demonstrate that the given plan is sub-optimal. If we define a new plan as ${\Pi^{*}}^i_j = \Pi^i_j - \Pi^j_i$ and ${\Pi^{*}}^j_i = 0$, we can verify that the resultant plan ${\Pi^{*}}$ incurs a lower transportation cost than $\Pi$. 
The preceding derivation establishes the tightness of our announcement, indicating that the optimal transport plan cannot be extended across the entire time interval.

In order to confirm the existence of such a scenario, we explicitly construct it in the case where $K=1$.
Since $P(t) = P(0) e^{Q_D t}$, we can obtain the analytical solution through eigen decomposition with difference equations, yielding the following outcomes: $\lambda_i = 2 cos(i \pi / S) - 2$ and $v_i = (1, cos(\theta_i/2), ... , cos((2S-1)\theta_i/2))$, where $\theta_i = \arccos((\lambda_i + 2) / 2)$. 
Subsequently, we can assess a basic scenario wherein $S = 3$ and $P(t = 0) = (0.1, 0, 0.9)$. It can be observed that $P_{0}(t = 0) > P_{1}(t = 0)$ and $P_{0}(t = 0.1) < P_{1}(t = 0.1)$.
However, the discussion presented above does not deny the existence of a long term optimal transport process. And from an application perspective, it is worth finding out a process with minimal uncertainty.

\subsection{Definition of probability flow on universal discrete process.}

In contrast to continuous processes, which necessitate the stochastic term to be a Brownian motion, there are few assumptions regarding the discrete stochastic term. As a result, the consideration of the drift term becomes unnecessary as it can be assimilated into the stochastic term. However, it is worth exploring the potential distinctive properties that may arise from treating these two terms separately. 

\subsection{Practical applications.}
Analogously to the effect of continuous probability flow on the continuous diffusion model, we believe that reducing uncertainty can also bring many benefits to discrete diffusion models. For instance, by selecting appropriate initial data, we can generate results that are pertinent to the initial data to attain controllable generation. Additionally, due to the excellent property of sampling certainty reduction, we can perform operations such as interpolation in latent code to complete data editing.

\subsection{Infinite horizon case in Proposition 2.}
The infinite horizon case is not addressed in our study due to the presence of singularities that pose significant challenges. For instance, in the case of Brownian motion, the distribution at $t = \infty$ assumes a uniform distribution over the entire ${\mathbb{R}^n}$, which is not well-defined. 

Additionally, the probability flow of Brownian motion at $t = 0$ also experiences a singularity. By taking the limit of the right-hand side of Eq. \ref{conti_prob_flow} and let $d_i = x_t - x_i$, we obtain:

\begin{equation}
\begin{aligned}
    \lim_{t \to 0^+} -\frac{1}{2} \nabla_{x_t} p_B (x_t, t) = & \lim_{t \to 0^+} \frac{\sum_i exp(-\frac{d_i^2}{2t}) \frac{d_i}{2t}}{\sum_j exp(-\frac{d_j^2}{2t})} \\
    = & \lim_{z \to +\infty} \sum_i \frac{d_i}{\sum_j exp((d_i^2 - d_j^2) z) / z}.
\end{aligned}
\end{equation}
Since
\begin{equation}
\lim_{z \to +\infty} exp((d_i^2 - d_j^2)z) / z = 
\begin{cases}
+\infty ~~~~ \text{if}~~d_i^2 > d_j^2, \\
0 ~~~~~~~~~ \text{if}~~d_i^2 \leq d_j^2,
\end{cases}
\end{equation}
we have
\begin{equation}
\lim_{t \to 0} - \frac{1}{2} \nabla_{x_t} p_B (x_t, t) 
= \lim_{z \to +\infty} \frac{d_{i_{min}}}{\sum_j exp((d_{i_{min}}^2 - d_j^2) z) / z},
\end{equation}

where $i_{min} = \mathop{\arg\min}\limits_{i} d_i^2 $. (noting that $i_{min}$ may not be unique, but we exclude this scenario as it does not significantly affect our analysis). Consequently, we obtain:

\begin{equation}
  \lim_{t \to 0^+} -\frac{1}{2} \nabla_{x_t} p_B (x_t, t)=\begin{cases}
    0, & \text{if}~~x_{t=0}=x_{i_{min}},\\
  d_{i_{min}} * \infty, & \text{else}.
  \end{cases}
\end{equation}

where $d_{i_{min}} * \infty$ indicates that the vector is oriented in the direction of $d_{i_{min}}$ and has an infinite norm. Consequently, the right-hand side of Eq. \ref{conti_prob_flow} lacks Lipschitz continuity, leading to non-unique solutions. Actually, if Eq. \ref{conti_prob_flow} has a unique solution near $t = 0$, the distribution $p_B(x,t)$ will always be a summation of Dirac deltas, which contradicts Eq. \ref{init_dist}. Due to our reliance on the solution of ODE, we are unable to analyze the behavior in the vicinity of $t = 0$. Consequently, we have limited our study to finite intervals. 

\end{document}